\documentclass[sn-basic]{sn-jnl}% Basic Springer Nature Reference Style/Chemistry Reference Style

\usepackage{multirow}%
\usepackage{amsmath,amssymb,amsfonts}%
\usepackage{amsthm}%
\usepackage{mathrsfs}%
\usepackage[title]{appendix}%
\usepackage{xcolor}%
\usepackage{textcomp}%
\usepackage{manyfoot}%
\usepackage{booktabs}%
\usepackage{algorithm}%
\usepackage{algorithmicx}%
\usepackage{algpseudocode}%
\usepackage{listings}%

\usepackage{anyfontsize} % Fixes "Font shape not available" warnings
\usepackage{type1cm}     % Allows smooth scaling of math fonts

\usepackage{amssymb,amsfonts,bm,mathtools}
\usepackage{mathrsfs}
\usepackage{dsfont}
\usepackage{bbm}
\usepackage{amsmath}

\usepackage{times}
\usepackage{microtype}
\usepackage[normalem]{ulem} % Fix: Removed hidden space
\usepackage{hyperref}
\usepackage{url}
\usepackage{pifont}
\usepackage[most]{tcolorbox}

\usepackage{booktabs}
\usepackage{multirow}
\usepackage[table]{xcolor}
\usepackage{tabularx}
\usepackage{graphicx}   % For \resizebox
\usepackage{array}      % For the >{\bfseries} column specifier
\usepackage{wrapfig}
\usepackage{colortbl}
\usepackage{enumitem}
\usepackage{subcaption}
\usepackage{tikz}
\usepackage{pgfplots}
\pgfplotsset{compat=1.18}

\usepackage{algorithm}
\usepackage{algorithmicx}
\usepackage{algpseudocode}

\definecolor{myGreen}{HTML}{33FF00}
\definecolor{myRed}{HTML}{FF3030}
\definecolor{cvprblue}{rgb}{0.21,0.49,0.74}
\definecolor{cvprgreen}{rgb}{0.2, 0.6, 0.0}
\definecolor{salmon}{RGB}{250,128,114}
\definecolor{lightgrayrow}{RGB}{248,248,248}

\definecolor{headerblue}{RGB}{232,240,255}
\definecolor{lightcyan}{RGB}{225,250,255}
\definecolor{yesgreen}{RGB}{40,160,70}
\definecolor{nored}{RGB}{210,50,50}

\newcommand{\cmark}{\textcolor{yesgreen}{\ding{51}}}
\newcommand{\xmark}{\textcolor{nored}{\ding{55}}}

\newcommand{\ourmodel}{\textsc{FoCUS}}

\theoremstyle{thmstyleone}%
\theoremstyle{thmstyletwo}%

\theoremstyle{thmstylethree}%

\begin{document}
\title[Article Title]{\ourmodel: Bridging Fine-Grained Recognition and Open-World Discovery across Domains}
%%=============================================================%%
%% Vaibhav Rathore (Corresponding Author)
%%=============================================================%%
\author*[1]{\fnm{Vaibhav} \sur{Rathore}}\email{vaibhav.rathor.in@gmail.com}

\author[2]{\fnm{Divyam} \sur{Gupta}}\email{divs25.iitb@gmail.com}

\author[3]{\fnm{Moloud} \sur{Abdar}}\email{m.abdar@uq.edu.au}

\author[1]{\fnm{Subhasis} \sur{Chaudhuri}}\email{sc@iitb.ac.in}

\author*[1]{\fnm{Biplab} \sur{Banerjee}}\email{bbanerjee@iitb.ac.in}

%%=============================================================%%
%% Affiliations
%%=============================================================%%
\affil*[1]{\orgdiv{Indian Institute of Technology Bombay}, \orgname{Mumbai}, \orgaddress{\country{India}}}

\affil*[2]{\orgdiv{Northeastern University}, \orgname{Boston}, \orgaddress{\country{USA}}}

\affil[3]{\orgdiv{The University of Queensland}, \orgname{Brisbane}, \orgaddress{\country{Australia}}}

\abstract{We introduce the first unified framework for \textit{Fine-Grained Domain-Generalized Generalized Category Discovery} (FG-DG-GCD), bringing open-world recognition closer to real-world deployment under domain shift. Unlike conventional GCD, which assumes labeled and unlabeled data come from the same distribution, DG-GCD learns only from labeled source data and must both recognize known classes and discover novel ones in unseen, unlabeled target domains. This problem is especially challenging in fine-grained settings, where subtle inter-class differences and large intra-class variation make domain generalization significantly harder.
To support systematic evaluation, we establish the first \textit{FG-DG-GCD benchmarks} by creating identity-preserving \textit{painting} and \textit{sketch} domains for CUB-200-2011, Stanford Cars, and FGVC-Aircraft using controlled diffusion-adapter stylization. On top of this ,we propose {\ourmodel}, a single-stage framework that combines \textit{Domain-Consistent Parts Discovery} (DCPD) for geometry-stable part reasoning with \textit{Uncertainty-Aware Feature Augmentation} (UFA) for confidence-calibrated feature regularization through uncertainty-guided perturbations.
Extensive experiments show that {\ourmodel} outperforms strong GCD, FG-GCD, and DG-GCD baselines by \textbf{3.28\%}, \textbf{9.68\%}, and \textbf{2.07\%}, respectively, in clustering accuracy on the proposed benchmarks. It also remains competitive on coarse-grained DG-GCD tasks while achieving nearly \textbf{3$\times$} higher computational efficiency than the current state of the art.  \footnote{Code and datasets will be released upon acceptance.}}
\keywords{Domain Generalization , Generalized Category Discovery , Fine-Grained Visual Categorization, Open-World Learning, Representation Learning}

\maketitle

\section{Introduction}

Deep visual recognition models~\cite{abbas2019comprehensive,elgendy2020deep} have achieved remarkable success in \emph{closed-world} settings~\cite{sem-survey,ssl-survey,sslssl}, where the label space is fixed and training and test samples are drawn from the same distribution. Real-world deployment, however, rarely satisfies either assumption. Practical systems must operate in \emph{open-world} environments~\cite{osr-survey,wen2024cross,ODG1}, where unseen categories can emerge after training and data distributions may shift substantially across domains. This problem is particularly acute in \emph{fine-grained recognition}~\cite{wei2021fine,xu2025deep}, where categories differ through subtle, localized attributes rather than coarse semantic cues. Under such conditions, even modest changes in texture, illumination, abstraction, or viewpoint can suppress the very evidence needed for discrimination, causing severe feature drift and unreliable transfer across domains. Applications such as biodiversity monitoring~\cite{biodiversity1}, industrial inspection~\cite{manufacturingda,manufacturingopen}, and surveillance or transportation analytics~\cite{airport1} therefore demand recognition systems that can \emph{generalize across domains} while also \emph{discovering previously unseen fine-grained categories}. We define the former as \emph{Fine-Grained Domain Generalization (FG-DG)}, focusing on robust known-class recognition under shift, and the latter as \emph{Generalized Category Discovery (GCD)}.

\begin{wrapfigure}[22]{r}{0.5\textwidth}
    \centering
    \vspace{-24pt}
    \includegraphics[width=1.0\linewidth]{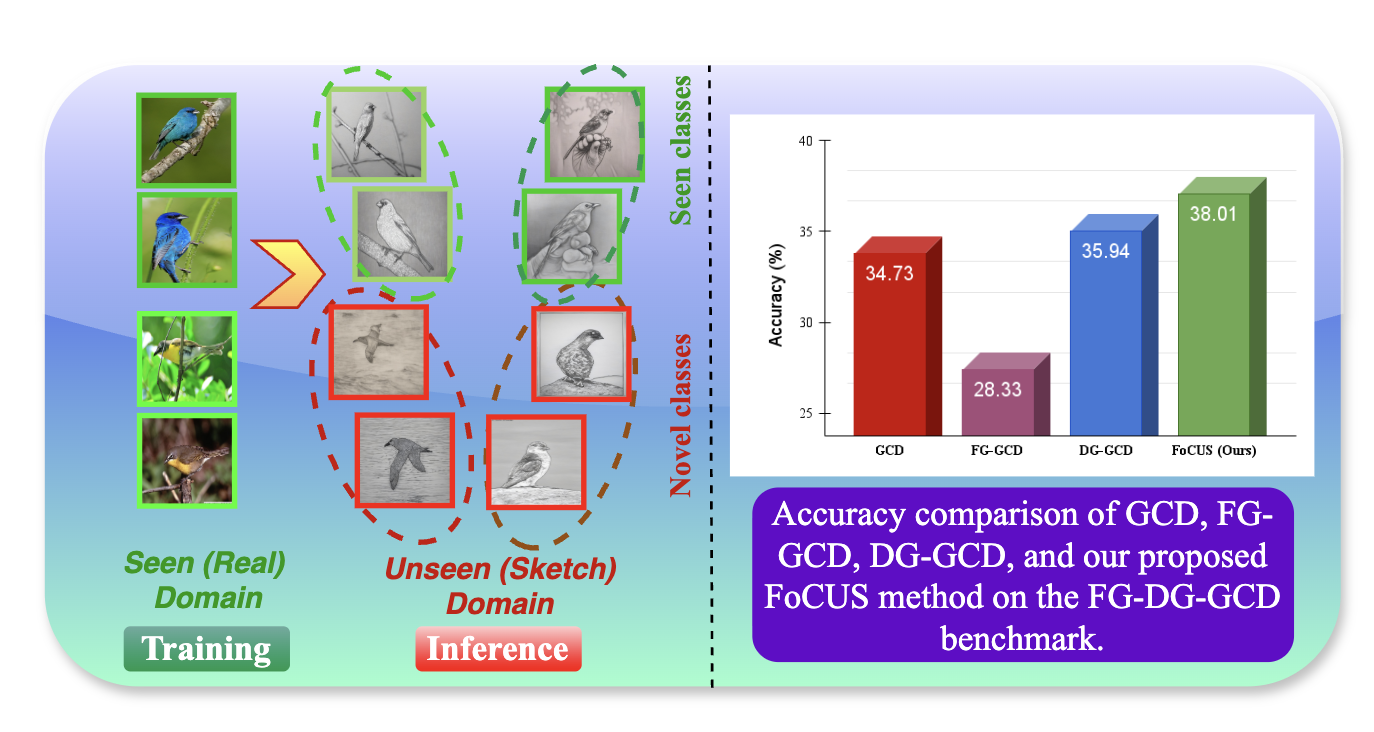}
    \captionsetup{font=small, justification=raggedright}
    \caption{
        \textbf{FG-DG-GCD} extends DG-GCD~\cite{dg2net} to the fine-grained regime, where a model trained on labeled source classes must simultaneously recognize known and discover novel categories in unseen, distribution-shifted domains. Compared with coarse-grained DG-GCD~\cite{dg2net,Rathore2025HiDISC}, this setting is substantially harder due to subtle inter-class cues, high intra-class similarity, and stronger sensitivity to domain-induced appearance changes. \ourmodel\ addresses this challenge through part-consistent representation learning and uncertainty-aware open-space regularization, outperforming existing GCD, FG-GCD, and DG-GCD baselines on the proposed benchmarks (Table~\ref{tab:results}).
    }
    \label{fig:teaser}
\end{wrapfigure}
\noindent

% \begin{figure}[t]
%     \centering
%     \includegraphics[width=1.0\linewidth]{images/teaser2.png}
%     % \vspace{-20pt}
%    \caption{
% \textbf{FG-DG-GCD} extends DG-GCD~\cite{dg2net} to the fine-grained regime, where a model trained on labeled source classes must simultaneously recognize known and discover novel categories in unseen, distribution-shifted domains. Compared with coarse-grained DG-GCD~\cite{dg2net,Rathore2025HiDISC}, this setting is substantially harder due to subtle inter-class cues, high intra-class similarity, and stronger sensitivity to domain-induced appearance changes. \ourmodel\ addresses this challenge through part-consistent representation learning and uncertainty-aware open-space regularization, outperforming existing GCD, FG-GCD, and DG-GCD baselines on the proposed benchmarks (Table~\ref{tab:results}).
% }
%     \label{fig:teaser}
%     % \vspace{-15pt}
% \end{figure}

Recent work on \emph{Domain-Generalized Generalized Category Discovery (DG-GCD)}~\cite{dg2net,Rathore2025HiDISC} moves toward this goal by combining \emph{Domain Generalization (DG)}~\cite{dg-survey1,dg-survey2}, which addresses distribution shift, with \emph{Generalized Category Discovery (GCD)}~\cite{gcd,gcd1,gcd2}, which jointly handles known and novel classes. In DG-GCD, a model is trained only on labeled source data, and must infer both seen and unseen categories in an unlabeled target domain whose distribution differs from the source. Existing DG-GCD methods rely on strategies such as diffusion-based domain synthesis, episodic meta-learning, or hyperbolic regularization~\cite{dg2net,Rathore2025HiDISC}. While these mechanisms are effective on coarse-grained benchmarks such as PACS~\cite{li2017deeper}, Office-Home~\cite{officehome}, and DomainNet~\cite{peng2019moment}, they are not designed for fine-grained transfer. In fine-grained settings, preserving local structure is critical: small parts, contours, and relative geometry often determine class identity (see Fig.\ref{fig:teaser}). Yet synthetic stylization can distort these cues, and global geometric regularization can suppress the localized evidence required for category discovery.

Conversely, fine-grained generalized category discovery (FG-GCD) methods~\cite{fgvc1,fgvc2,cvprapl,rastegar2023learn} improve recognition by leveraging part discovery, token selection, or region-aware transformers. However, these methods typically assume that labeled and unlabeled data share the same distribution, and thus do not address domain shift. Many also entangle part localization with clustering or rely on target-domain access during training, which violates the DG-GCD setting. More fundamentally, they often exploit appearance-dependent cues such as color and texture, which are highly domain-variant, while lacking mechanisms to isolate geometry-stable evidence or calibrate predictions in open-set regions. As a result, existing coarse-grained DG-GCD methods do not preserve fine-grained locality, and existing fine-grained GCD methods do not generalize across domains.
% \vspace{-10pt}

\begin{table}[t]
\centering
\renewcommand{\arraystretch}{1.25}
\setlength{\tabcolsep}{6pt}
\resizebox{\linewidth}{!}{
\begin{tabular}{lcccccc}
\toprule
\rowcolor{headerblue}
\textbf{Method} & \textbf{Fine-} & \textbf{Domain} & \textbf{Novel Class} & \multicolumn{2}{c}{\textbf{Design Choice}} & \textbf{Part-} \\
\rowcolor{headerblue}
 & \textbf{Grained} & \textbf{Shift} & \textbf{Discovery} & \textbf{Synthetic-Free} & \textbf{Episodic-Free} & \textbf{Aware} \\
\midrule
\rowcolor{lightgrayrow} DG~\cite{closedDG1,ODG1,SDG3} & \xmark & \cmark & \xmark & \cmark & \cmark & \xmark \\
\rowcolor{lightgrayrow} FG-DG~\cite{FSDG,csfg} & \cmark & \cmark & \xmark & \cmark & \cmark & \cmark \\
\rowcolor{white} (FG)GCD~\cite{gcd,gcd1,gcd2} & \cmark & \xmark & \cmark & \cmark & \cmark & \xmark \\
\rowcolor{lightgrayrow} DG-GCD~\cite{dg2net,Rathore2025HiDISC} & \xmark & \cmark & \cmark & \xmark & \xmark & \xmark \\
\midrule
\rowcolor{lightcyan} \textbf{{\ourmodel} (Ours)} & \textbf{\cmark} & \textbf{\cmark} & \textbf{\cmark} & \textbf{\cmark} & \textbf{\cmark} & \textbf{\cmark} \\
\bottomrule
\end{tabular}}
\captionsetup{font=small}
\caption{{\ourmodel} is the first framework to jointly address \textbf{fine-grained recognition}, \textbf{domain generalization}, and \textbf{novel class discovery} within a single formulation. Unlike prior DG-GCD methods~\cite{dg2net,Rathore2025HiDISC}, it is \emph{synthetic-free} and \emph{episodic-free}, reducing training FLOPs by \textbf{3$\times$} and \textbf{300$\times$} compared with \textsc{HiDISC}~\cite{Rathore2025HiDISC} and \textsc{DG$^2$CD-Net}~\cite{dg2net}, respectively, while delivering stronger generalization and discovery performance.}
\label{tab:teaser_supple}
\vspace{-22pt}
\end{table}

These limitations expose a clear research gap, summarized in Table~\ref{tab:teaser_supple}: \emph{there is currently no framework that jointly addresses fine-grained discrimination, domain generalization, and open-world novel class discovery}. We formalize this missing setting as \textbf{Fine-Grained Domain-Generalized Generalized Category Discovery (FG-DG-GCD)}. Importantly, FG-DG-GCD is not a simple combination of FG-DG and GCD. It introduces a new dual challenge. First, the representation must remain sufficiently precise to distinguish known fine-grained categories under domain shift. Second, it must avoid collapsing novel target samples into the known-class manifold, thereby preserving the structural flexibility required for discovery. This makes FG-DG-GCD fundamentally harder than closed-set fine-grained DG, since the model must learn invariances that remove domain-specific variation without erasing the subtle category-defining cues needed to separate both seen and unseen classes.

A second major bottleneck is the absence of suitable benchmarks. Standard fine-grained datasets such as CUB-200-2011~\cite{CUB_200_2011}, Stanford Cars~\cite{scars}, and FGVC-Aircraft~\cite{fgvc_aircarft} contain only natural-image domains and therefore cannot evaluate domain-generalized discovery. Although diffusion- or GAN-based stylization~\cite{yu2024analysis,ge2024tuning} can generate auxiliary domains, naïve style transfer often alters shape, erases discriminative parts, or introduces artifacts that corrupt category identity. To enable rigorous study of FG-DG-GCD, we construct the first dedicated benchmarks by generating identity-preserving \texttt{painting} and \texttt{sketch} domains using a controlled SDXL-adapter pipeline~\cite{podell2023sdxl}. By conditioning the stylization process on structure and contour cues, we retain semantic identity and part geometry while inducing substantial shifts in texture, tone, and abstraction. This yields a reproducible evaluation protocol that stresses both domain robustness and fine-grained novel-category discovery.

\noindent\textbf{Our Approach.}
To address FG-DG-GCD, we propose {\ourmodel} (\textbf{F}ine-grained \textbf{O}pen-world \textbf{C}ategory discovery with \textbf{U}ncertainty and part\textbf{S}), a single-stage, synthetic-free, and episodic-free framework that learns solely from labeled source data. Our central premise is that effective FG-DG-GCD requires solving \emph{two coupled problems}: (1) preserving domain-stable fine-grained evidence, and (2) preventing overconfident collapse in open-space regions where novel classes emerge. {\ourmodel} addresses these challenges through two complementary modules.

% \textbf{(i) Domain-Consistent Parts Discovery (DCPD).}
\textbf{(i) Domain-Consistent Parts Discovery (DCPD).}
A primary limitation of current DG-GCD frameworks is their reliance on predominantly global representations, which often fail to capture the subtle local variations required for fine-grained transfer under domain shift. While recent fine-grained GCD methods like \textsc{APL}~\cite{cvprapl} attempt to leverage part-based attention, they are optimized for domain-homogeneous settings and utilize \textit{globally-shared queries} that frequently succumb to source-domain texture bias. Consequently, such methods remain susceptible to overfitting on stylistic attributes (e.g., color or texture) that do not generalize across domains. 
DCPD is designed to resolve this gap. Unlike \textsc{APL}, it introduces \emph{image-conditioned part queries} that are dynamically grounded in a \emph{geometry-stable ViT attention prior}. This enables the model to discover semantic parts through invariant structural evidence---such as contours and object layout---rather than domain-variant appearance. Furthermore, instead of relying on the rigid, hand-crafted diversity constraints or global templates found in prior work, DCPD promotes part distinctness through an \emph{adaptive, competitive patch-to-part routing} mechanism. This architectural bias ensures that discovered parts remain spatially coherent and semantically transferable, providing stable anchors for both known-class recognition and novel-class discovery across distribution-shifted domains.

% A major weakness of existing DG-GCD methods is that they learn predominantly global representations, which are insufficient for fine-grained transfer under domain shift. At the same time, existing part-learning methods often rely on static latent embeddings, target-dependent learning, or auxiliary diversity objectives that do not explicitly promote domain consistency. DCPD is designed to resolve this gap. It introduces \emph{image-conditioned part queries} that attend over a geometry-stable ViT attention prior, allowing the model to discover semantic parts using structural evidence rather than domain-variant texture. This is a key novelty of our design: part discovery is explicitly biased toward contours and object layout, which are more stable across domains and more informative for fine-grained category identity. Moreover, DCPD avoids hand-crafted diversity losses. Instead, part distinctness emerges through a \emph{competitive, annealed patch-to-part routing} mechanism that discourages redundant part assignments and promotes coherent specialization. This produces cleaner, more transferable part representations for both known-class recognition and novel-class grouping.

\textbf{(ii) Uncertainty-Aware Feature Augmentation (UFA).}
Even with better part representations, FG-DG-GCD remains vulnerable to a second failure mode: novel target samples can be absorbed into overconfident known-class regions, degrading both calibration and discovery. Existing DG-GCD approaches lack an explicit mechanism to regularize such open-space uncertainty. UFA addresses this deficiency by modeling uncertainty directly in the embedding space. It introduces a \emph{tripartite OOD sampling strategy} that synthesizes feature perturbations corresponding to near-class uncertainty, between-class ambiguity, and far-away unknown regions. These perturbations are used to regularize the classifier toward high-entropy predictions away from well-supported source manifolds. The result is a better-calibrated representation space in which known classes remain compact and discriminative, while novel classes are less likely to collapse into existing decision regions. Thus, UFA complements DCPD by improving not only robustness but also the structural separability needed for open-world category discovery.

Together, DCPD and UFA define a unified framework that is specifically tailored to FG-DG-GCD: DCPD preserves \emph{what} should remain invariant across domains in fine-grained recognition, while UFA regulates \emph{where} the model should remain uncertain to support novel-class emergence. Unlike prior DG-GCD approaches, \ourmodel\ does not depend on expensive domain synthesis or episodic training; unlike prior fine-grained GCD methods, it does not assume source--target distribution overlap or target-domain access during training.

In summary, our contributions are three-fold:
\begin{itemize}
    \item We formalize \textbf{Fine-Grained Domain-Generalized Generalized Category Discovery (FG-DG-GCD)}, a new and practically relevant open-world recognition problem that unifies fine-grained discrimination, domain generalization, and novel-class discovery.
    
    \item We establish the \textbf{first FG-DG-GCD benchmarks} by constructing identity-preserving \texttt{painting} and \texttt{sketch} domains for CUB-200-2011, Stanford Cars, and FGVC-Aircraft via controlled SDXL-adapter stylization, enabling systematic evaluation under fine-grained domain shift.
    
    \item We propose \textbf{{\ourmodel}}, the first unified, synthetic-free, and episodic-free framework for FG-DG-GCD, featuring \textbf{Domain-Consistent Parts Discovery (DCPD)} for geometry-stable fine-grained representation learning and \textbf{Uncertainty-Aware Feature Augmentation (UFA)} for calibrated open-space regularization. Extensive experiments show that \ourmodel\ sets a strong new baseline for both fine-grained DG-GCD and transfer to coarse-grained DG-GCD benchmarks.
\end{itemize}
\section{Paper Organization}
The remainder of this paper is organized as follows. 

\begin{itemize}
    \item In Section~\ref{sec:related_works}, we provide a comprehensive review of literature concerning domain generalization, fine-grained recognition, and generalized category discovery. 

    \item Section~\ref{sec:data} describes the construction of our identity-preserving fine-grained benchmarks and the SDXL-based synthetic domain generation pipeline. 

    \item Section~\ref{sec:method} formally defines the \textbf{FG-DG-GCD} problem and Section~\ref{sec:methodology} provide details of our \ourmodel\ framework, focusing on the \textbf{Domain-Consistent Parts Discovery (DCPD)} and \textbf{Uncertainty-aware Feature Augmentation (UFA)} modules.

    \item In Section~\ref{sec:experiments}, we present the experimental setup, implementation details, and comparison against state-of-the-art baselines. 

    \item Section~\ref{sec:ablations} provides an extensive ablation analysis of our method’s components and hyperparameter sensitivity. 

    \item Section~\ref{sec:conclusions} concludes the paper and discusses future research directions.

\end{itemize}

\section{Related Works}
\label{sec:related_works}

\noindent\textbf{(a) Domain Generalization.}
Domain Generalization (DG)~\cite{dg-survey1,dg-survey2} aims to learn models from labeled source domains that generalize to unseen target distributions. Classical multi-source DG methods~\cite{closedDG1,closedDG3,closedDG4,closedDG2,closedDG5,closedDG6} exploit inter-domain diversity to learn domain-invariant representations, whereas single-source DG methods~\cite{SDG1,SDG2,SDG3,SDG4} compensate for the lack of source diversity through stronger regularization, augmentation, or adversarial objectives. Open-set DG~\cite{ODG1,ODG2,ODG3,ODG4} further relaxes the closed-set assumption by allowing unseen categories at test time. However, these methods typically group all unseen samples into a single \emph{unknown} category and do not attempt to separate them into semantically meaningful classes. In contrast, our FG-DG-GCD setting requires the model not only to remain robust under domain shift, but also to \emph{discover and cluster multiple distinct novel categories} in unseen target domains without any target-domain access during training.

\noindent\textbf{(b) Fine-Grained Domain Generalization.}
Recent fine-grained domain generalization (FG-DG) methods~\cite{FSDG,csfg} focus on learning domain-invariant cues for visually similar categories, often by emphasizing subtle local patterns and fine structural details. While these methods are effective in closed-set recognition, they are not designed for open-world discovery. In particular, they assume that all test samples belong to the known training label space, and therefore lack mechanisms to preserve latent separability for emerging unseen categories. As a result, directly extending FG-DG methods to FG-DG-GCD can lead to \emph{representation collapse}, where novel classes are inadvertently absorbed into known-class regions. Compared with standard FG-DG, FG-DG-GCD introduces a substantially harder dual objective: the model must simultaneously maintain high precision for known fine-grained classes and preserve sufficient structure in the embedding space to enable novel-category discovery under domain shift.

\noindent\textbf{(c) (Fine-Grained) Category Discovery.}
Category discovery has evolved from Novel Category Discovery (NCD)~\cite{ncd1,ncd2,ncd3,ncd4,ncd5,hsu2017learning}, where only unlabeled novel classes are inferred, to Generalized Category Discovery (GCD)~\cite{gcd1,gcd2,gcd3,gcd4,gcd5,gcd6,gcd7}, where known and novel categories are handled jointly. Existing GCD methods typically rely on clustering~\cite{gcd1}, pseudo-labeling~\cite{gcd2}, consistency regularization, or distillation~\cite{gcd3}, and generally assume that labeled and unlabeled data arise from the same or closely aligned distributions. More recent fine-grained NCD/GCD methods~\cite{fgvc1,fgvc2,cvprapl,fgvc4,fgvc5,rastegar2023learn} improve discovery through part-based reasoning, token selection, or region-aware representations. However, these methods remain essentially intra-domain: they do not explicitly address domain shift, and their reliance on appearance-sensitive cues makes them fragile when subtle category-defining evidence changes across styles or domains. Consequently, they are not suitable for the FG-DG-GCD setting, where fine-grained discrimination and novel-category discovery must both remain reliable under unseen distribution shift.

\noindent\textbf{(d) Category Discovery Across Domains.}
A related line of work studies category discovery across domains~\cite{rongali2024cdadnetbridgingdomaingaps,wang2024exclusivestyleremovalcross,wen2024cross,wang2024hilolearningframeworkgeneralized}, often assuming access to unlabeled target-domain samples during training. Such methods can adapt their representations using target information and therefore serve as upper bounds rather than true domain-generalized solutions. Domain-Generalized Generalized Category Discovery (DG-GCD)~\cite{dg2net,Rathore2025HiDISC} removes this assumption and instead tackles category discovery in unseen target domains without target supervision. DG$^2$CD-Net~\cite{dg2net} relies on diffusion-based domain synthesis and episodic task optimization, while \textsc{HiDISC}~\cite{Rathore2025HiDISC} introduces tangent-space augmentation and hyperbolic regularization. Although effective on coarse-grained benchmarks, these approaches are not tailored to fine-grained recognition, where preserving local structure and subtle part geometry is essential. In such settings, aggressive stylization may distort discriminative details, while global hierarchical priors may oversmooth or compress the feature geometry needed to separate visually similar categories. This motivates a different design for FG-DG-GCD, one that preserves fine-grained locality while remaining robust to domain shift.

\noindent\textbf{(e) Part-Based Representation Learning.}
Part-based modeling improves recognition by decomposing objects into semantically meaningful local substructures. Early approaches relied on part annotations, pose supervision, or attention maps~\cite{wei2021fine,huang2016part,lin2015deep,zhang2014part}, whereas more recent self-supervised or weakly supervised methods~\cite{chen2019looks,donnelly2022deformable,huang2020interpretable,wang2021interpretable,Zhang_2020} learn part prototypes automatically from visual correspondences and foreground structure~\cite{shrack2025pairwisematchingintermediaterepresentations,cvprapl}. Adaptive Part Learning (APL)~\cite{cvprapl}, for instance, employs DINO-guided queries~\cite{dino} to discover semantically aligned regions across images. Despite their success, existing part-based fine-grained recognition and discovery methods, including PartGCD~\cite{fgvc1}, SelEx~\cite{fgvc2}, and InfoSieve~\cite{rastegar2023learn}, typically assume access to unlabeled target data or couple part discovery with downstream clustering. These assumptions are incompatible with DG-GCD, where the target domain is entirely unseen during training. Our work instead uses part learning as a source-only inductive bias for discovering geometry-stable, domain-consistent local cues that can support both fine-grained transfer and open-world discovery.

\section{Datasets and Synthetic Domain Construction}
\label{sec:data}

\definecolor{boxblue}{RGB}{232,240,255}
\definecolor{boxgreen}{RGB}{235,250,240}
\definecolor{titleblue}{RGB}{0,51,102}

We evaluate the proposed \textbf{FG-DG-GCD} setting on three standard fine-grained recognition benchmarks:
\textbf{CUB-200-2011}~\cite{CUB_200_2011} (200 bird species),
\textbf{Stanford Cars}~\cite{scars} (196 car models), and
\textbf{FGVC-Aircraft}~\cite{fgvc_aircarft} (100 aircraft variants).

These datasets contain only a single natural-image domain, which prevents direct evaluation of domain-generalized category discovery. To enable cross-domain evaluation under the FG-DG-GCD setting, we construct two additional stylized domains that preserve object structure while altering global appearance.

Specifically, for each dataset we generate three domains:
\texttt{real} (original images), \texttt{painting}, and \texttt{sketch}.
The stylized domains introduce substantial shifts in texture, tone, and abstraction while maintaining semantic identity and spatial composition.

Representative examples and additional visualizations are provided in the \textbf{Sup. Mat}.

\subsection{\textbf{SDXL Adapter Pipeline for Domain Synthesis}}
\label{subsec:sdxl}

To construct the \texttt{painting} and \texttt{sketch} domains, we generate stylized versions of each real image $\mathbf{x}$ using a \textbf{Stable Diffusion XL (SDXL)}~\cite{podell2023sdxl} adapter pipeline conditioned on a \textbf{line-art structural signal}. The key objective is to induce substantial domain shift in appearance while preserving geometric layout, object structure, and fine-grained identity cues. This is crucial in our setting, since naive stylization can easily distort the subtle local attributes that define fine-grained classes.

\resizebox{0.85\linewidth}{!}{% % Wrap the entire environment
\begin{tcolorbox}[
    colback=boxblue,
    colframe=titleblue,
    title=\textbf{Technical Configuration},
    arc=2pt,
    boxrule=0.8pt
]
\noindent\textbf{Control Signal.}
A high-level structural map $\mathbf{c}$ is extracted from the input image using the \texttt{ControlNet Auxiliary Lineart} detector:
\[
\begin{aligned}
\mathbf{c} \leftarrow \mathrm{LineartDetector}(&\texttt{det\_res}=384,
&\texttt{img\_res}=1024).
\end{aligned}
\]
\noindent\textbf{Backbone and Adapter.}
We employ the \textbf{SDXL base model} (\texttt{stabilityai/\allowbreak sd-xl-base-1.0}) together with the \textbf{T2I Adapter for line art} \texttt{TencentARC/\allowbreak t2i-adapter-\allowbreak lineart-\allowbreak sdxl-1.0}~\cite{mou2023t2i}. The generation pipeline uses half-precision inference, \textit{xFormers} optimization, and an \textbf{Euler Ancestral} scheduler.
\end{tcolorbox}}

\paragraph{\textbf{Domain-Specific Prompts.}}
To modify style while preserving semantic content and object layout, we use domain-specific prompts designed to alter appearance without changing category identity:

\begin{itemize}[leftmargin=*, labelsep=5pt]
    \item \textbf{\textcolor{titleblue}{Painting:}} ``A canvas painting of the same scene, painted with oils or acrylics, rich textures, visible brushstrokes, natural color palette, same composition, same objects.''
    \item \textbf{\textcolor{titleblue}{Sketch:}} ``A quickdraw-style black-and-white line-art sketch of the same scene, minimal detail, same composition, same objects.''
\end{itemize}

\begin{tcolorbox}[colback=white, colframe=gray!50, title=\textit{Negative Constraints}, fonttitle=\bfseries, arc=0pt, boxrule=0.5pt]
\small
\textbf{Painting:} Photorealism, digital art, vector graphics, flat colors, sharp lines, exaggerated details. \\
\textbf{Sketch:} Realism, detailed textures, shading, color, painterly style, complex backgrounds.
\end{tcolorbox}

\paragraph{\textbf{Identity Preservation.}}
The use of a line-art control map ensures that the generated domains largely preserve object geometry, silhouette, and spatial arrangement while varying domain-specific appearance factors such as texture, shading, brush patterns, and abstraction. In practice, this helps retain fine-grained discriminative attributes---for example, bird beak shape and head markings, aircraft wing and tail structure, or the grille and headlight configuration of cars---that are essential for category identity.

\begin{wrapfigure}[10]{r}{0.40\textwidth}
\centering
\vspace{-35pt} % Aligns the top of the box with the adjacent text line
\begin{tcolorbox}[colback=boxgreen, colframe=black!70, arc=3pt, boxrule=1pt]
\centering
\small
\textbf{Fixed Generation Hyperparameters}

% \vspace{5pt}

\begin{tabular}{l|l}
\toprule
\textbf{Parameter} & \textbf{Value} \\
\midrule
Sampling Steps   & 30 \\
Guidance Scale   & 7.5 \\
Adapter Scale    & 1.0 \\
Resolution       & 1024 px \\
Precision        & FP16 \\
Optimization     & xFormers \\
\bottomrule
\end{tabular}
\end{tcolorbox}
\end{wrapfigure}

\subsection{\textbf{Reproducibility and Quality Control}}
\label{subsec:repro}

To ensure reproducibility, all synthetic images are generated using fixed hyperparameters across datasets and domains.

These controlled settings allow the synthetic domains to be generated consistently across all classes and datasets, reducing variance introduced by the generation process itself. More importantly, they ensure that the induced domain shift primarily reflects stylistic transformation rather than uncontrolled semantic corruption.

% \noindent\textbf{Synthetic Domain Construction.}
\subsection{\textbf{Validation of Synthetic Domain}s}
New style domains are synthesized using Stable Diffusion XL (SDXL) with a \textbf{T2I-Adapter} conditioned on \emph{line-art maps}, preserving geometry and object identity while altering texture, tone, and abstraction.
This produces semantically faithful yet perceptually distinct \texttt{painting} and \texttt{sketch} domains, validated by high FID scores (Table~\ref{tab:dataset_overview}) and manual inspection.
% Visualizations are provided in the Sec:.\ref{sec:data}.
% New style domains are synthesized using Stable Diffusion XL (SDXL) guided by a \textbf{T2I-Adapter} conditioned on \emph{line-art maps} from source images, preserving geometry and object identity while varying texture, tone, and abstraction.
% This yields semantically faithful yet perceptually distinct \texttt{painting} and \texttt{sketch} domains, verified via high FID~\cite{eiter1994computing} scores (Table~\ref{tab:dataset_overview}) as well as manually.
% Further details and visualizations are provided in the \textbf{Sup. Mat.}

\begin{table}[!ht]
\centering
\setlength{\tabcolsep}{4.5pt}
\renewcommand{\arraystretch}{1.1}
\resizebox{0.97\columnwidth}{!}{%
\begin{tabular}{lcccccccccc}
\toprule
\multirow{2}{*}{\textbf{Dataset}} & 
\multirow{2}{*}{\textbf{\#Classes}} & 
\multicolumn{1}{c}{\textbf{Split}} &
\multicolumn{3}{c}{\textbf{\#Images}} &
\multirow{2}{*}{\textbf{\#Domains}} &
\multicolumn{3}{c}{\textbf{FID Scores}~($\downarrow$)} \\
\cmidrule(lr){3-3} \cmidrule(lr){4-6} \cmidrule(lr){8-10}
 &  & \textbf{K:N} & \textbf{Real} & \textbf{Painting} & \textbf{Sketch} &  & \textbf{R$\rightarrow$P} & \textbf{R$\rightarrow$S} & \textbf{S$\rightarrow$P} \\
\midrule
\rowcolor{lightgrayrow}CUB-200-2011-MD  & 200 & 100:100 & 11{,}788 & 11{,}788 & 11{,}788 & 3 & 55.13 & 160.18 & 106.95 \\
\rowcolor{white}Stanford Cars-MD & 196 & 98:98   & 16{,}365 & 16{,}365 & 16{,}365 & 3 & 29.62 & 62.86  & 41.63 \\
\rowcolor{lightgrayrow}FGVC-Aircraft-MD & 100 & 50:50   & 10{,}000 & 10{,}000 & 10{,}000 & 3 & 64.33 & 170.27 & 81.78 \\
\bottomrule
\end{tabular}}
\captionsetup{font=small}
\caption{\textbf{Summary of the tri-domain fine-grained benchmarks} used in this study. Each dataset comprises \texttt{real} photographs and their \texttt{painting} and \texttt{sketch} counterparts synthesized via SDXL with T2I-Adapter. FID~\cite{eiter1994computing} scores measure cross-domain shifts, reflecting the realism and diversity of the generated domains.}
\vspace{-15pt}
\label{tab:dataset_overview}
\end{table}

\paragraph{LPIPS Analysis of Synthetic Domains}
% \label{subsec:lpips}
To further validate the perceptual consistency of our painting and sketch domains, we report LPIPS distances computed using AlexNet and VGG backbones. Across all datasets (Table~\ref{tab:lpips}), sketch domains exhibit moderately lower distances than painting domains, indicating that sketches preserve more structural information relative to paintings. The distances remain within the typical range observed in cross-style translation tasks, confirming that our SDXL + T2I-Adapter pipeline produces style-shifted yet semantically faithful variants. These LPIPS scores complement the FID evaluations, jointly demonstrating that our synthetic domains maintain high perceptual quality.

\begin{table}[ht]
\centering
\renewcommand{\arraystretch}{1.2}
\setlength{\tabcolsep}{8pt}
\resizebox{0.8\textwidth}{!}{%
\begin{tabular}{lcccc}
\toprule
\textbf{Dataset} & \textbf{Domain Pair} & \textbf{LPIPS (AlexNet)} & \textbf{LPIPS (VGG)} \\
\midrule
\multirow{2}{*}{CUB-200-2011-MD} 
 & Real $\rightarrow$ Sketch   & 0.5019 & 0.5558 \\
 & Real $\rightarrow$ Painting & 0.5564 & 0.5801 \\
\midrule
\multirow{2}{*}{FGVC-Aircraft-MD} 
 & Real $\rightarrow$ Sketch   & 0.4525 & 0.5441 \\
 & Real $\rightarrow$ Painting & 0.5158 & 0.5585 \\
\midrule
\multirow{2}{*}{Stanford Cars-MD} 
 & Real $\rightarrow$ Sketch   & 0.4284 & 0.4718 \\
 & Real $\rightarrow$ Painting & 0.4599 & 0.4784 \\
\bottomrule
\end{tabular}%
}
\captionsetup{font=small}
\caption{\textbf{LPIPS distances between real and synthetic domains.}  
Lower values indicate higher perceptual similarity. Scores are computed with AlexNet and VGG backbones.}
\label{tab:lpips}
\end{table}
\paragraph{\textbf{Ethical Considerations and Data Usage.}}
The construction of the multi-domain benchmarks (CUB-200-2011-MD, Stanford Cars-MD, and FGVC-Aircraft-MD) adheres to the following ethical guidelines:

\begin{itemize}[leftmargin=*, labelsep=5pt]
    \item \textbf{Dataset Licensing:} The source datasets—CUB-200-2011 \cite{CUB_200_2011}, Stanford Cars~\cite{scars}, and FGVC-Aircraft~\cite{fgvc_aircarft}—are used strictly for non-commercial research purposes in accordance with their original licenses. Our stylization process does not claim ownership over the underlying semantic content.
    \item \textbf{Generative Safety:} We employ Stable Diffusion XL (SDXL) solely for \textbf{stylistic domain transformation}. By using a fixed adapter-based pipeline with controlled structural prompts, we ensure that the generated content remains semantically faithful to the source and does not introduce malicious or hallucinated artifacts.
    \item \textbf{Privacy and Bias:} No human identities, faces, or sensitive personal attributes are processed or generated in these benchmarks. While generative models can inherit biases from their training data, our prompts are designed to be category-neutral (focusing on ``painting'' or ``sketch'' styles), thereby minimizing the risk of introducing social or cultural biases into the fine-grained recognition task.
    \item \textbf{Intended Impact:} These benchmarks are intended to facilitate research in robust, open-world recognition and category discovery. We advocate for the responsible use of these datasets to improve the reliability of computer vision systems in critical applications like biodiversity monitoring and industrial inspection.
\end{itemize}
\section{Problem Definition}
\label{sec:method}

We adopt the DG-GCD formulation~\cite{dg2net}, where training occurs on a single labeled \emph{source domain} $\mathcal{D}_s = \{(\mathbf{x}_i^s, y_i^s)\}_{i=1}^{N_s}$, with label set $\mathcal{Y}_{\mathrm{known}}$ of size $|\mathcal{Y}_{\mathrm{known}}|$. At test time, the model is evaluated on an unseen \emph{target domain} $\mathcal{D}_t = \{\mathbf{x}_j^t\}_{j=1}^{N_t}$ drawn from a shifted distribution $\mathcal{P}_t(\mathbf{x}) \neq \mathcal{P}_s(\mathbf{x})$. The target label space extends to $\mathcal{Y}_t = \mathcal{Y}_{\mathrm{known}} \cup \mathcal{Y}_{\mathrm{novel}}$, where $\mathcal{Y}_{\mathrm{novel}} \cap \mathcal{Y}_{\mathrm{known}} = \emptyset$. The goal is to (i) induce clusters for target samples from $\mathcal{Y}_{\mathrm{known}}$ that correspond to the known categories, and (ii) \emph{discover and cluster} target samples from $\mathcal{Y}_{\mathrm{novel}}$ into $|\mathcal{Y}_{\mathrm{novel}}|$ groups, without access to any labeled or unlabeled target data during training. In the FG-DG-GCD setting, classes within $\mathcal{Y}_{\mathrm{known}}$ and $\mathcal{Y}_{\mathrm{novel}}$ differ by subtle, localized cues, while intra-class variability (due to pose, texture, or viewpoint) often matches or exceeds inter-class separation, i.e., $\sigma^2_{\text{intra}} \gtrsim \Delta_{\text{inter}}$. This small-margin regime makes both feature clustering and domain generalization significantly challenging.

\section{Proposed Methodology: \ourmodel}
\label{sec:methodology}

\begin{figure}[t]
    \centering
    \includegraphics[width=\linewidth]{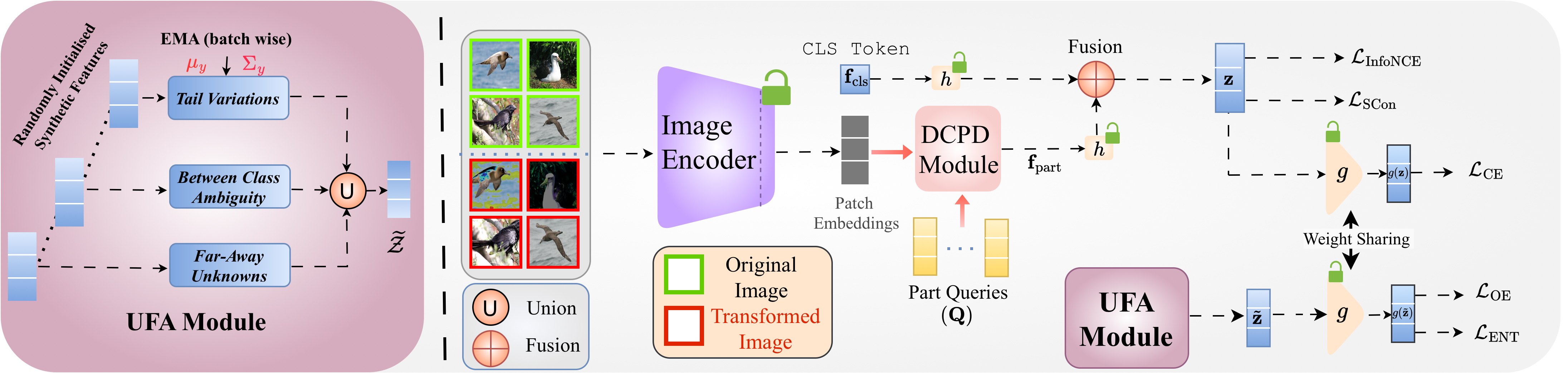}
    \captionsetup{font=small}
    \caption{\textbf{Overview of \ourmodel.} A ViT encoder produces global \texttt{[CLS]} and patch-level features. The \textbf{Domain-Consistent Parts Discovery (DCPD)} module extracts geometry-stable part features $\mathbf{f}_{\text{part}}$, which are fused with the global representation $\mathbf{f}_{\text{cls}}$ and optimized using InfoNCE ($\mathcal{L}_{\text{InfoNCE}}$) and supervised contrastive ($\mathcal{L}_{\text{SCon}}$) losses for fine-grained discrimination. In parallel, the \textbf{Uncertainty-Aware Feature Augmentation (UFA)} module uses EMA-based class statistics $(\boldsymbol{\mu}_y, \boldsymbol{\Sigma}_y)$ to synthesize outlier features $\widetilde{\mathbf{\mathcal{Z}}}$ from class-tail, between-class, and hypersphere regions. A shared classifier $g(\cdot)$ is trained on real features using cross-entropy, and on synthetic outliers using energy-based outlier exposure ($\mathcal{L}_{\text{OE}}$) and entropy maximization ($\mathcal{L}_{\text{ENT}}$). Together, these two pathways yield discriminative, robust, and calibrated embeddings for category discovery under unseen domain shift.}
    % \vspace{-5pt}
    \label{fig:model_architecture}
\end{figure}

We aim to address the coupled challenges of fine-grained discrimination, domain shift, and open-set uncertainty in a unified manner.  
To this end, we introduce {\ourmodel}, a simple yet powerful framework built upon two complementary modules:  
DCPD (Sec. \ref{sec:apl1}) for robust representation learning, and  
UFA (Sec. \ref{sec:ufa}) for calibrated open-space regularization.  
An overview of the architecture is shown in Fig.~\ref{fig:model_architecture}.

%%%%%%%%%%%%%%%%%%%%%%%%%%%%%%%%
%%%%%%%%%%%%%%%%%%%%%%%%%%%%%%%%

\subsection{\textbf{Domain-Consistent Parts Discovery}}
\label{sec:apl1}

\begin{figure}[!ht]
    \centering
    % --- Left Side: The Image ---
    \begin{minipage}[c]{0.49\textwidth}
        \centering
        \includegraphics[width=0.75\linewidth]{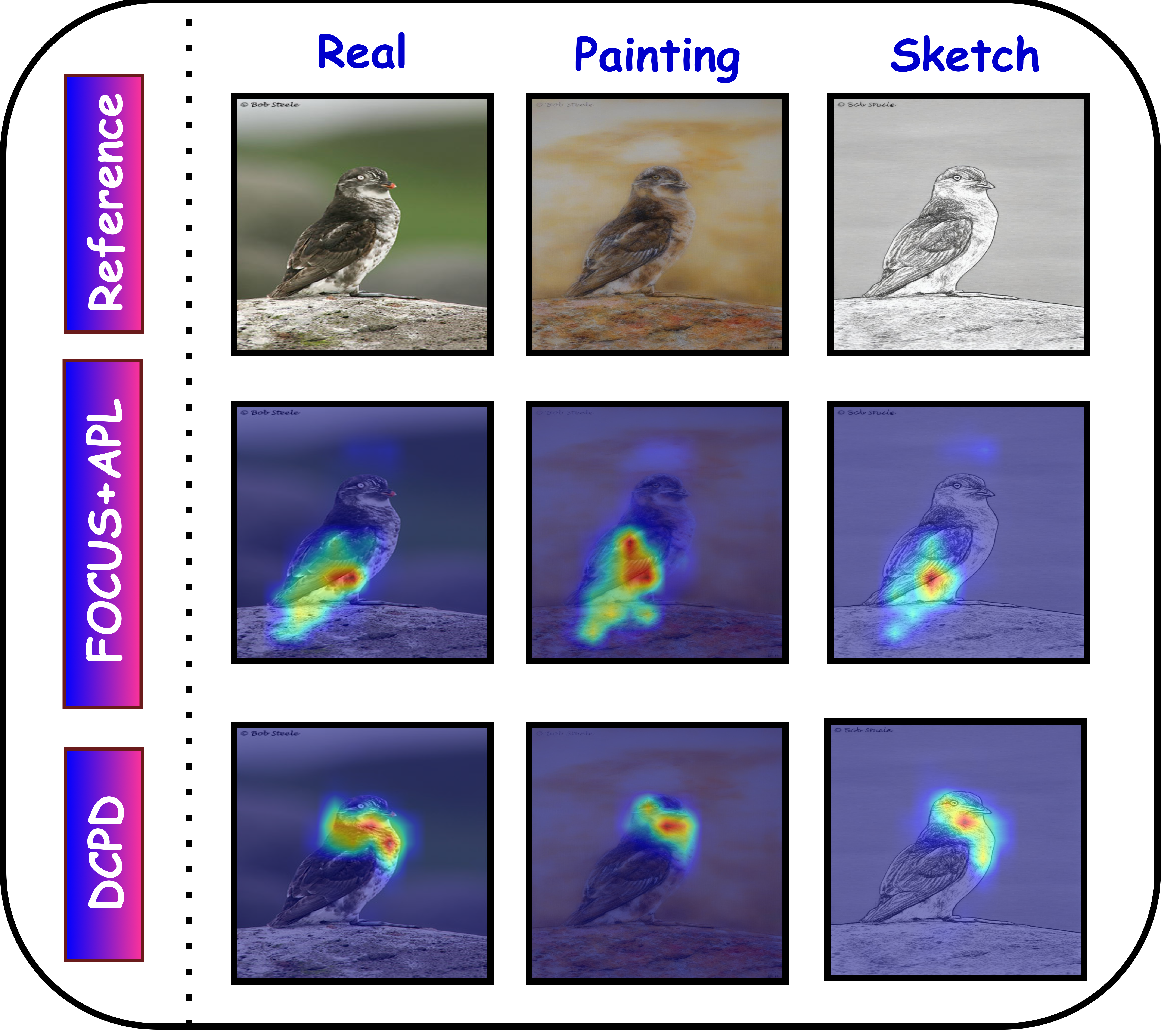}
    \end{minipage}
    \hfill
    % --- Right Side: The Caption ---
    \begin{minipage}[c]{0.48\textwidth}
    \captionsetup{font=small, justification=raggedright}
    \captionof{figure}{\textbf{Domain-Consistent Attention across Domains.} 
    Compared to prior part-localization methods (FOCUS+APL \cite{cvprapl}), our DCPD module maintains stable attention on the same geometric regions (e.g., bird’s head) across domain shifts from \textit{Real} $\rightarrow$ \textit{Painting} $\rightarrow$ \textit{Sketch}, demonstrating strong domain-invariant part discovery. See \textbf{Sup. Mat.} for KL divergence between the cross-domain attention maps for quantitative validation.}
    \label{fig:dcpd_attention_maps}
    \end{minipage}
\end{figure}

Standard ViTs~\cite{vit} encode an image as a sequence of patch embeddings that are globally aggregated via a \texttt{[CLS]} token.  
While this global summarization captures holistic semantics, it tends to suppress the subtle local variations crucial for fine-grained recognition.  
Under domain shifts (e.g., real $\rightarrow$ sketch or painting), superficial cues such as color or texture change significantly, whereas geometric regularities—such as the \emph{curvature of a beak} or the \emph{contour of a wheel}—remain largely invariant.  
The proposed {DCPD} module exploits this geometric stability to identify and encode structure-preserving parts that remain consistent across domains (Fig.\ref{fig:dcpd_attention_maps}).  
These parts serve as reliable anchors for visual reasoning, providing interpretable and transferable features even when domain appearance changes drastically.

\noindent

Among part-based methods~\cite{fgvc1,fgvc2,rastegar2023learn}, APL~\cite{cvprapl} represents a significant precursor that leverages DINO~\cite{dino} self-attention as a prior for unsupervised part localization. While APL utilizes a Gumbel-Softmax formulation to achieve hard patch-to-part assignment, its reliance on a set of \textbf{globally shared, learnable queries} makes it susceptible to \textbf{distribution shifts}. Specifically, because these queries are optimized across the entire dataset without explicit domain conditioning, they often capture texture-biased features that lack robustness when encountering novel domains---such as the transition from natural images to sketches or paintings. 
In contrast, \textbf{DCPD} introduces an explicit \textit{domain-aware architectural bias} to ensure part-level stability across varying distributions. DCPD advances beyond the static query paradigm by: 
(i) \textit{dynamically conditioning} part queries on each image's intrinsic geometric structure rather than relying on domain-agnostic global priors, 
(ii) enforcing a \textit{stricter structural exclusivity} to prevent semantic drift and maintain sharp part boundaries under heavy stylistic distortion, and (iii) systematically \textit{integrating localized part cues with the global scene context}. 
This design yields domain-invariant, part-aware representations that serve as stable anchors for visual reasoning, facilitating robust category discovery without the need for manual part annotations or domain-specific labels.
(Refer to the \textbf{Sup. Mat.} for further comparison between APL and the proposed DCPD.)
\vspace{1em}
% {r} aligns it to the right. {0.45\textwidth} sets the width of the wrap box.
% \begin{wrapfigure}{R}{0.45\textwidth}
%     \centering
%     \vspace{-15pt} % Adjust spacing to align with text
%     \includegraphics[width=\linewidth]{images/DCPD_Attention.png}
%     \caption{\textbf{Domain-Consistent Attention across Domains.} 
% Compared to prior part-localization methods (FOCUS+APL \cite{cvprapl}), our DCPD module maintains stable attention on the same geometric regions (e.g., bird’s head) across domain shifts from \textit{Real} $\rightarrow$ \textit{Painting} $\rightarrow$ \textit{Sketch}, demonstrating strong domain-invariant part discovery. See \textbf{Sup. Mat.} for KL divergence between the cross-domain attention maps for quantitative validation.}
%     \label{fig:dcpd_attention_maps}
%     \vspace{-10pt}
% \end{wrapfigure}
%%%%%%%%%%%%%%%%%%%%%%%%%%%%%%%%

Let an input image $\mathbf{x}$ be divided into $N$ non-overlapping patches, each mapped to a $D$-dimensional embedding by a pretrained ViT encoder (e.g., DINO), forming  
$\mathbf{F}_{\text{patch}}\!\in\!\mathbb{R}^{N\times D}$.  
DCPD refines these patch features through four interlinked stages, progressively evolving from coarse geometric attention to semantically fused representations.

\noindent \textbf{(1) Self-Attention Priors.}  
We first extract coarse, geometry-relevant priors from the ViT’s emergent \texttt{[CLS]}$\rightarrow$patch attention. Let $\mathbf{\mathcal{A}}\in\mathbb{R}^{H\times (N+1)\times (N+1)}$ denote the multi-head self-attention tensor. Unlike existing methods that utilize raw attention features, we interpret these maps as a \textit{topological mask} to filter out non-structural noise. For each head $h$, the attention weight $A_{\text{cls}}^{(h,n)}$ is thresholded to select geometry-salient patches:
\begin{equation}
\label{eq:threshold}
M_{h,n} = \mathbbm{1}\!\big[A_{\text{cls}}^{(h,n)} \ge \epsilon_h(\rho)\big],
\end{equation}
where we set $\rho = 0.30$. The coarse prior $\mathbf{f}_{\text{prior}}^{(h)}$ is obtained via masked mean pooling:
\begin{equation}
\label{eq:prior}
\mathbf{f}_{\text{prior}}^{(h)} = \frac{\sum_n M_{h,n}\,\mathbf{F}_{\text{patch}}^{(n)}}{\sum_n M_{h,n} + \delta}.
\end{equation}
These priors $\mathbf{F}_{\text{prior}}\in\mathbb{R}^{H\times D}$ provide a spatial blueprint of geometry-stable regions (e.g., contours) that remain invariant across domain stylizations.

\noindent \textbf{(2) Image-Conditioned Part Queries.}  
To move beyond the limitations of globally-shared queries~\cite{cvprapl} which can succumb to source-domain texture bias, DCPD introduces \textit{instance-adaptive conditioning}. We initialize $T$ learnable part prototypes $\mathbf{Q}\in\mathbb{R}^{T\times D}$ representing abstract structural concepts. These are dynamically transformed into image-specific queries $\mathbf{Q}_I$ via cross-attention:
\begin{equation}
\mathbf{Q}_I = \mathrm{MHA}_{\text{cross}}(\mathbf{Q},\,\mathbf{F}_{\text{prior}},\,\mathbf{F}_{\text{prior}}).
\end{equation}
By grounding $\mathbf{Q}$ in the instance-specific $\mathbf{F}_{\text{prior}}$, we ensure that the resulting part detectors are driven by the geometric configuration of the current image rather than an averaged global template.

\noindent \textbf{(3) Differentiable Patch-to-Part Assignment.}  
To resolve the assignment of $N$ patches to $T$ discovered parts while maintaining end-to-end differentiability, we employ a Straight-Through Gumbel-Softmax relaxation. The assignment probability $\mathcal{H}_{n,t}$ is computed using cosine similarity $\mathbf{S}_{n,t}$ and Gumbel noise $\mathbf{g}_{n,t}$:
\begin{equation}
\mathcal{H}_{n,t} = \frac{\exp((\mathbf{S}_{n,t}+\mathbf{g}_{n,t})/\tau)}{\sum_{t'} \exp((\mathbf{S}_{n,t'}+\mathbf{g}_{n,t'})/\tau)}.
\end{equation}
As $\tau \rightarrow 0$, $\mathcal{H}_{n,t}$ enforces a near-exclusive mapping. This structural exclusivity is critical for Fine-Grained GCD, as it prevents the "feature bleeding" often seen in soft-attention mechanisms, thereby preserving the distinctness of subtle part variations.

\noindent \textbf{(4) Part--Global Fusion.}  
Unlike previous approaches that discard the global token in favor of aggregated part features~\cite{cvprapl}, DCPD maintains a \textit{dual-stream} representation to harmonize localized structural cues with holistic object context. Each discovered part embedding $\mathbf{p}_t$ aggregates its assigned patches:
\begin{equation}
\mathbf{p}_t = \frac{\sum_n \mathcal{H}_{n,t}\mathbf{F}_{\text{patch}}^{(n)}}{\sum_n \mathcal{H}_{n,t}+\delta}.
\end{equation}
The resulting descriptors $\mathbf{P} = [\mathbf{p}_1,\dots,\mathbf{p}_T]$ are averaged into a unified part-based representation $\mathbf{f}_{\text{part}} = \frac{1}{T}\sum_t \mathbf{p}_t$. Finally, we fuse this with the global \texttt{[CLS]} feature $\mathbf{f}_{\text{cls}}$ through a non-linear projection head $h(\cdot)$:
\begin{equation}
\mathbf{z} = \mathrm{L2Norm}\! \big(h(\mathbf{f}_{\text{cls}}) + h(\mathbf{f}_{\text{part}})\big).
\label{eq:fusion}
\end{equation}
This fusion strategy ensures that $\mathbf{z}$ inherits the high-level category semantics of $\mathbf{f}_{\text{cls}}$ while being anchored by the domain-stable geometric precision of $\mathbf{f}_{\text{part}}$. By avoiding the total replacement of the global token, DCPD achieves a more resilient embedding that preserves discriminability even when part-level cues are partially degraded by extreme domain shifts.

\begin{figure}[!ht]
    \centering
    % --- Left Side: The Image ---
    \begin{minipage}[c]{0.48\textwidth}
        \centering
        \includegraphics[width=0.85\linewidth]{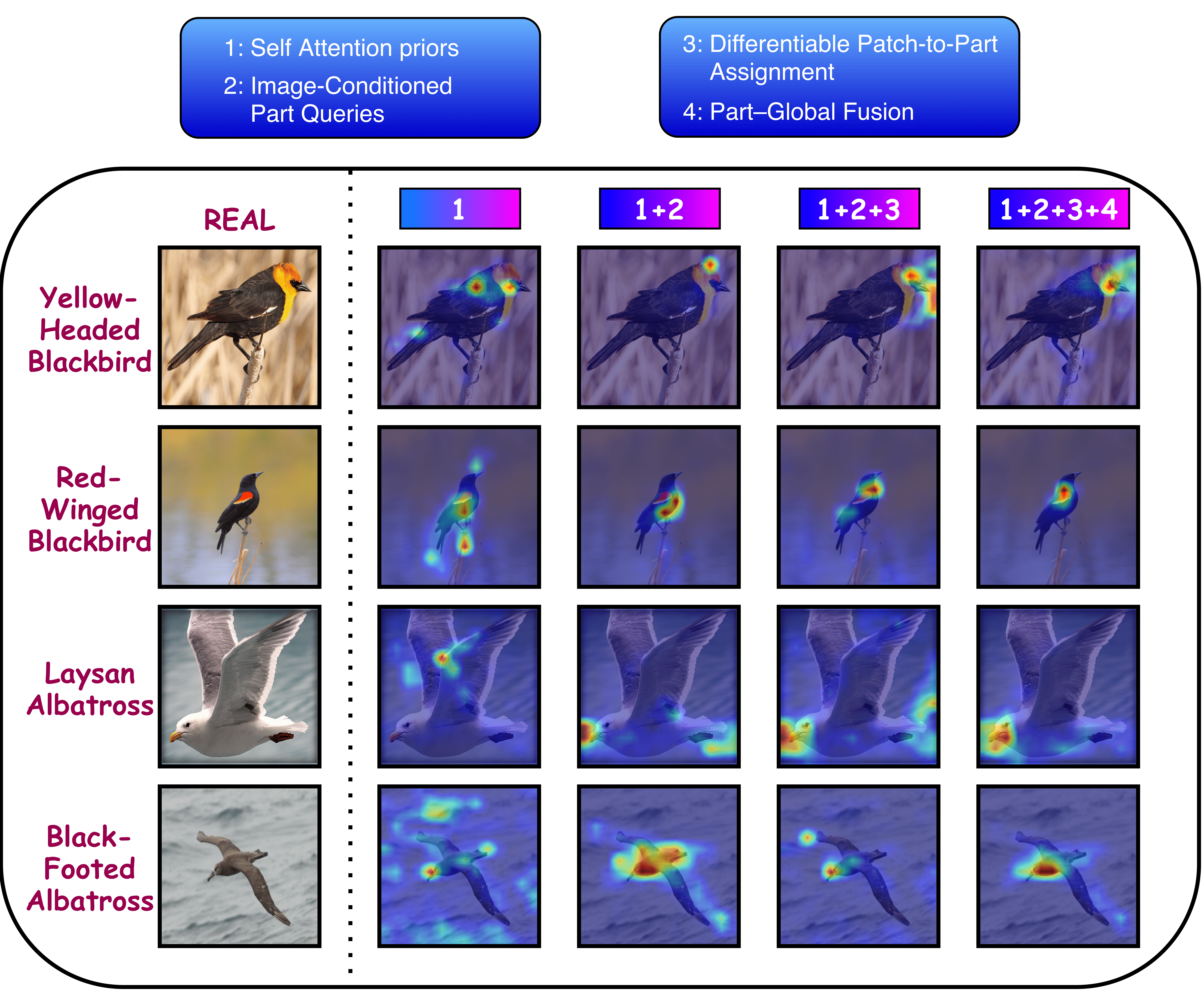}
    \end{minipage}
    \hfill
    % --- Right Side: The Caption ---
    \begin{minipage}[c]{0.50\textwidth}
    \captionsetup{font=small, justification=raggedright}
        \caption{
            \textbf{Progressive Attention Refinement within DCPD}. Left to Right,the evolution of : {(1) Self-Attention Priors} provide initial coarse localization; {(2) Image-Conditioned Part Queries} refine attention to align with instance-specific geometries; {(3) Differentiable Patch-to-Part Assignment} creates sharp, spatially exclusive part delineations; and finally, {(4) Part-Global Fusion} integrates these precise structural cues into a unified representation.
        }
        \label{fig:attention_maps}
    \end{minipage}
\end{figure}

Each stage in DCPD progressively transforms raw patch embeddings into semantically meaningful, geometry-grounded representations (Fig. \ref{fig:attention_maps}).  
Self-Attention Priors identify stable structural cues, Image-Conditioned Queries adapt to instance geometry, the Gumbel-based Assignment enforces localized part exclusivity, and Part–Global Fusion harmonizes structure with semantics.  
By grounding representation learning in geometry rather than texture, DCPD consistently produces domain-invariant, interpretable, and fine-grained features.

% \begin{figure}[t]
%     \centering
%     \resizebox{0.90\columnwidth}{!}{
%     \includegraphics[width=0.85\linewidth]{images/FG_DGCD-Page-12.png}}
%     \caption{\textbf{Progressive Attention Refinement within DCPD}. Left to Right,the evolution of : {(1) Self-Attention Priors} provide initial coarse localization; {(2) Image-Conditioned Part Queries} refine attention to align with instance-specific geometries; {(3) Differentiable Patch-to-Part Assignment} creates sharp, spatially exclusive part delineations; and finally, {(4) Part-Global Fusion} integrates these precise structural cues into a unified representation.}
%     \label{fig:attention_maps}
%     % \vspace{-15pt}
% \end{figure}

\subsection{\textbf{UFA for Open-Space Calibration}}
\label{sec:ufa}

While DCPD yields domain-stable, geometry-grounded embeddings, it cannot by itself prevent overconfident predictions in sparse or unseen regions. Prior DG or open-set methods rely on image-level augmentations, style transfers, or uniform confidence regularization~\cite{zhou2021domain,hendrycks2018deep,kendall2017uncertainties,lakshminarayanan2017simple}, which either distort semantic geometry or overlook sample-specific uncertainty, leading to \emph{false absorption} of novel samples into known manifolds. Under the small-margin regime ($\sigma^2_{\text{intra}} \gtrsim \Delta_{\text{inter}}$), such indiscriminate sharpening miscalibrates decision boundaries and entangles feature clusters. UFA addresses this by modeling and regularizing uncertainty \emph{within the embedding space} through distribution-aware perturbations around known manifolds, enforcing adaptive, semantically consistent boundaries. Unlike geometry-altering stylization or diffusion augmentations, UFA operates directly on DCPD embeddings—producing robust, interpretable, and well-calibrated open-space recognition. Details follow.

\noindent\textbf{(1) Estimating Known-Class Statistics.}  
Given DCPD embeddings $\mathbf{z}$, we maintain class-conditional statistics using exponential moving averages (EMA).  
For each known class $y \in \mathcal{Y}_{\text{known}}$:
\begin{equation}
\begin{aligned}
\boldsymbol{\mu}_y \!\leftarrow\! \alpha_1\,\boldsymbol{\mu}_y 
+ (1-\alpha_1)\,\widehat{\mathbf{z}}_y, \\
\boldsymbol{\Sigma}_y \!\leftarrow\! 
\alpha_2\,\boldsymbol{\Sigma}_y + (1-\alpha_2)\,\widehat{\boldsymbol{\Sigma}}_y,
\end{aligned}
\end{equation}
where $\widehat{\mathbf{z}}_y$ and $\widehat{\boldsymbol{\Sigma}}_y$ denote the batch mean and covariance of class $y$, and $\alpha_{1/2}$ represent the EMA decay.  
These running estimates approximate each class’s local density, supporting controlled uncertainty sampling around known manifolds.

\begin{wrapfigure}[17]{r}{0.5\textwidth}
    \centering
    \vspace{-16pt} % Adjust to align top with text
\includegraphics[width=0.5\textwidth]{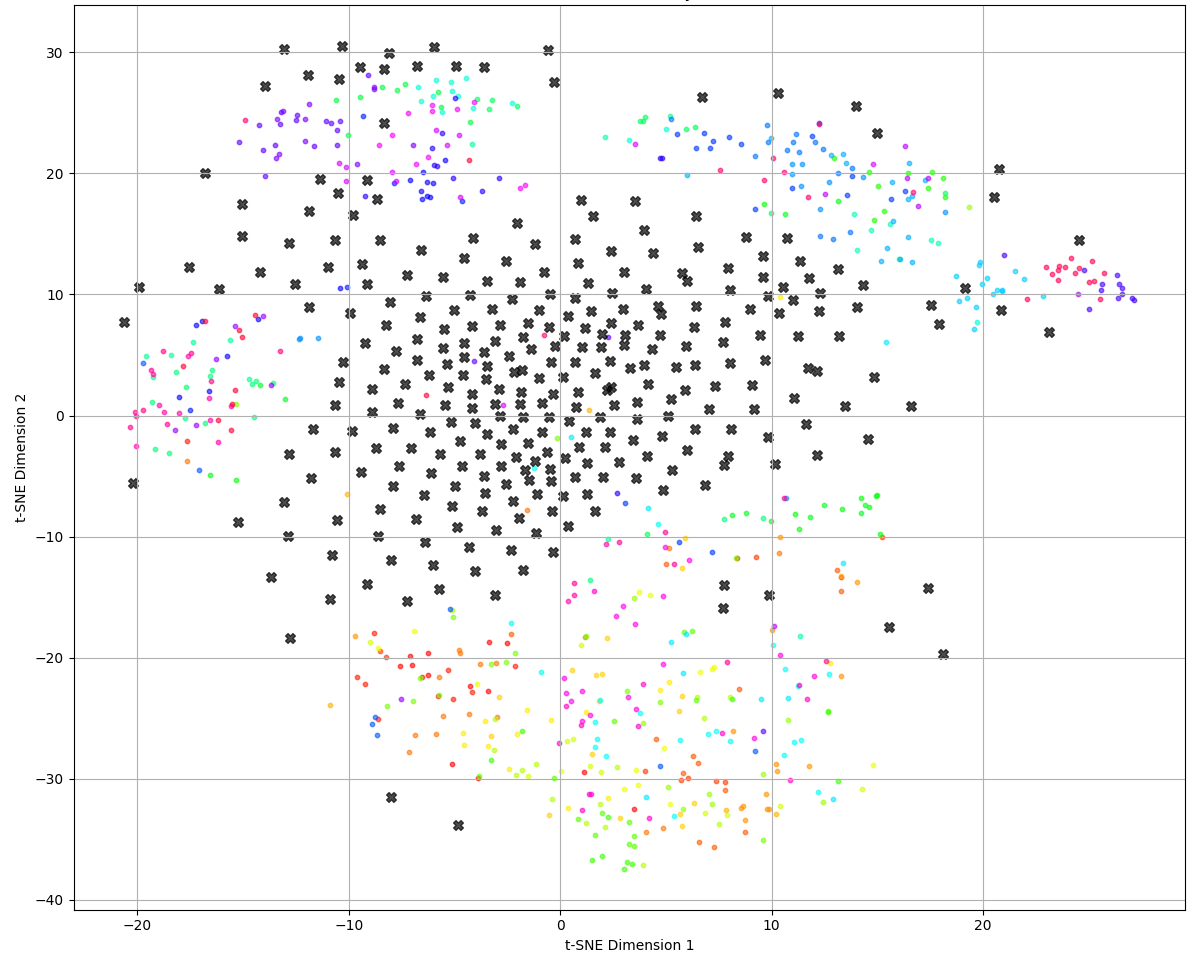}
    \caption{
    t-SNE~\cite{van2008visualizing} of \textbf{DCPD embeddings} with source (colored) and UFA outliers (black) occupying low-density boundaries, modeling open-space uncertainty and improving calibration.
    }
    \label{fig:UFA_outlier_tsne}
\end{wrapfigure}

\noindent\textbf{(2) Probabilistic OOD Proposals.} 
To simulate open-space uncertainty without the overhead of image-level generation, UFA samples synthetic feature proposals $\widetilde{\mathcal{Z}}$ directly within the DCPD embedding space. The effectiveness of this approach stems from the \textbf{synergy of three complementary sampling strategies}, each designed to provide structured coverage of the open space by targeting specific ``leaks'' in the decision boundary (Fig.~\ref{fig:UFA_outlier_tsne}):

\begin{itemize}[leftmargin=*, labelsep=5pt]
    \item \textbf{Tail-Variation Sampling (Near-OOD):} For each known class $y \in \mathcal{Y}_{\text{known}}$, we sample $\tilde{\mathbf{z}} \sim \mathcal{N}(\boldsymbol{\mu}_y, \beta\boldsymbol{\Sigma}_y)$, where $\beta > 1$ (typically $1.5 \text{--} 3.0$) inflates the class-conditional covariance. This shifts the probability mass into low-density regions surrounding the class manifold, mimicking rare, borderline instances or subtle domain drifts. By capturing these near-boundary variations, this sampler \emph{curbs local overconfidence} and forces the classifier to maintain sharp boundaries even under feature-space expansion.
    
    \item \textbf{Between-Class Ambiguity Mixing (Structured OOD):} To model semantic ambiguity in anisotropic regions between class centroids, we form convex mixtures of class means: $\tilde{\mathbf{z}} = \sum_{l=1}^{k} w_l \boldsymbol{\mu}_{y_l} + \boldsymbol{\epsilon}$, where $\mathbf{w} \sim \mathrm{Dirichlet}(\mathbf{1}_k)$ and $\boldsymbol{\epsilon} \sim \mathcal{N}(\mathbf{0}, \sigma^2 \mathbf{I})$. Sampling $\mathbf{w}$ from a uniform Dirichlet populates the transitional ``open'' zones between fine-grained manifolds. This encourages the model to treat ambiguous, high-overlap regions as non-identity space, preventing the false absorption of novel samples into the overlapping manifolds of known classes.
    
    \item \textbf{Far-Away Unknown Sampling (Far-OOD):} To generate directionally novel, out-of-support features, we sample $\mathbf{v} \sim \mathcal{N}(\mathbf{0}, \mathbf{I})$ and project to the unit hypersphere via $\tilde{\mathbf{z}} = \mathbf{v}/\|\mathbf{v}\|_2$. Normalizing these random Gaussian vectors produces features maximally separated in cosine space from known-class embeddings. These far-OOD proposals \emph{prevent open-space extrapolation} into distant, empty regions of the hypersphere, providing robust anchors for the energy and entropy regularization terms.
\end{itemize}

Our ablation studies (Tab.~\ref{tab:ablation_all}(a)) confirm that these strategies are strictly complementary; using any two in isolation leaves an incomplete OOD set, resulting in a \emph{leaky} boundary that significantly degrades novelty discovery. Together, they enable efficient, scalable OOD modeling that preserves the geometric integrity of the fine-grained embeddings while ensuring calibrated discovery under domain shift.

\noindent\textbf{(3) Open-Set Regularization.}  
We employ a cosine classifier of the form : 
% $ g_y(\mathbf{z})=\gamma\,\frac{\mathbf{w}_y^{\!\top}\mathbf{z}}{\|\mathbf{w}_y\|\|\mathbf{z}\|},$
% \vspace{-3pt}
\begin{equation}
g_y(\mathbf{z})=\gamma\,\frac{\mathbf{w}_y^{\!\top}\mathbf{z}}{\|\mathbf{w}_y\|\|\mathbf{z}\|},
\end{equation}
\noindent where $\mathbf{w}_y$ is the weight vector for class $y\!\in\!\mathcal{Y}_{\text{known}}$ and $\gamma$ is a learnable scaling factor.  
This formulation projects normalized embeddings onto a hyperspherical manifold, making decisions purely by angular alignment.  
Such geometry encourages compact intra-class clusters and large inter-class margins, improving separability in fine-grained, open-set conditions. However, even with angular normalization, the classifier may remain overconfident in sparse regions.To counter this, UFA introduces two complementary regularizers over the uncertainty samples $\tilde{\mathbf{z}}\!\in\!\widetilde{\mathbf{\mathcal{Z}}}$.

\textbf{\textit{(a) Energy-Based Outlier Exposure.}  }
We define the free energy:
\begin{equation}
% \vspace{-2pt}
E(\tilde{\mathbf{z}})=-\tau_{\text{temp}}\log\!\sum_{y\in\mathcal{Y}_{\text{known}}}\!\exp\!\big(g_y(\tilde{\mathbf{z}})/\tau_{\text{temp}}\big),
\end{equation}
where $\tau_{\text{temp}}$ controls temperature scaling.  
A margin ($m$) constraint ensures OOD samples remain high in energy:
\begin{equation}
\mathcal{L}_{\text{OE}}=\tfrac{1}{|\widetilde{\mathbf{\mathcal{Z}}}|}\!
\sum_{\tilde{\mathbf{z}}\in\widetilde{\mathbf{\mathcal{Z}}}}\!
\max(0,\,m-E(\tilde{\mathbf{z}})),
\label{eq:loss_oe}
\end{equation}
penalizing low-energy (overconfident) responses for open-space features (see Table \ref{tab:ablation_all} (b)).

\textbf{\textit{(b) Entropy Maximization.}  }
Energy separation alone may not suppress peaked softmax probabilities.  
We thus maximize predictive entropy on OOD samples:
% {\scriptsize
\begin{equation}
\begin{aligned}
p(y|\tilde{\mathbf{z}}) &=
\frac{\exp\!\big(g_y(\tilde{\mathbf{z}})/\tau_{\text{temp}}\big)}
     {\sum_{y'\in\mathcal{Y}_{\text{known}}}\exp\!\big(g_{y'}(\tilde{\mathbf{z}})/\tau_{\text{temp}}\big)}, \\[4pt]
\mathcal{L}_{\text{ENT}} &=
\tfrac{1}{|\widetilde{\mathbf{\mathcal{Z}}}|}
\sum_{\tilde{\mathbf{z}}\in\widetilde{\mathbf{\mathcal{Z}}}}
\sum_{y\in\mathcal{Y}_{\text{known}}}
- p(y|\tilde{\mathbf{z}})\log p(y|\tilde{\mathbf{z}}).
\end{aligned}
\end{equation}
% }

This promotes uniform predictions (high entropy) in open-space regions, further mitigating overconfidence.

Finally, the overall uncertainty regularization objective is defined as:
\begin{equation}
% \vspace{-1pt}
\mathcal{L}_{\text{UFA}}=\lambda_{\mathrm{OE}}\mathcal{L}_{\text{OE}}-\lambda_{\mathrm{ENT}}\mathcal{L}_{\text{ENT}},
% \vspace{-5pt}
\end{equation}
where $\lambda_{\mathrm{OE}}$ balances margin enforcement and $\lambda_{\mathrm{ENT}}$ entropy smoothing.  
Jointly, these terms reshape the classifier’s energy surface—keeping known-class clusters compact and confident while assigning high-energy, high-entropy responses to unseen samples, yielding calibrated open-space recognition under domain shift (see Table \ref{tab:ufa_comparison_accuracy}).

\noindent Detailed pseudocode is provided in \textbf{Sup. Mat.}

\subsection{Theoretical Superiority of UFA for Open-Space Calibration}
\label{sec:ufa_theory}

UFA improves open-space calibration by explicitly modeling uncertainty in the
DCPD embedding space. Its superiority can be established via three
complementary theoretical arguments: margin preservation, density-driven
regularization, and calibrated energy separation.

\subsubsection{Margin Preservation Under Domain Shift.}
Let $\mathcal{M}_y$ denote the DCPD manifold of class $y$, with clean
inter-class margin
\begin{equation}
\Delta_{\mathrm{inter}}
=
\min_{y\neq y'} 
\min_{\mathbf{z}\in\mathcal{M}_y,\mathbf{z}'\in\mathcal{M}_{y'}}
d(\mathbf{z},\mathbf{z}').
\end{equation}
Image-level augmentations or stylization methods modify $\mathbf{z}$ directly,
injecting perturbations orthogonal to the manifold and reducing the effective
margin to $\Delta_{\mathrm{inter}}' < \Delta_{\mathrm{inter}}$.

UFA perturbs only through the estimated class covariance
$\Sigma_y$, ensuring
\begin{equation}
\tilde{\mathbf{z}}\in 
\mathcal{N}(\mu_y,\beta\Sigma_y)
\quad \Rightarrow \quad
\tilde{\mathbf{z}} \in 
\mathrm{Tube}_\epsilon(\mathcal{M}_y),
\end{equation}
a controlled $\epsilon$-neighborhood of the manifold.  
This guarantees that intra-class variation increases without collapsing margins:
\begin{equation}
\Delta_{\mathrm{inter}}'
\approx
\Delta_{\mathrm{inter}}
\quad\text{(first-order invariant)}.
\end{equation}
Thus UFA preserves geometric separability better than OE, MixUp, CutMix, or
diffusion-based proposals.

\subsubsection{Density-Driven Open-Space Modeling.}
Let $p_y(\mathbf{z})$ denote the class density.  
UFA samples uncertainty points according to
\begin{equation}
\tilde{\mathbf{z}} \sim 
\underbrace{\mathcal{N}(\mu_y,\beta\Sigma_y)}_{\text{low-density tails}}
\;\cup\;
\underbrace{\mathrm{Dir}(\mu_{y_1},\dots)}_{\text{between-class}}
\;\cup\;
\underbrace{\mathrm{Uniform}(\mathbb{S}^{d-1})}_{\text{far-OOD}},
\end{equation}
which span the set
\begin{equation}
\mathcal{O}
=
\{\mathbf{z}: p_y(\mathbf{z}) < \tau\},
\end{equation}
the formal definition of open space.

In contrast, methods using image augmentations fail to guarantee that augmented
samples fall within $\mathcal{O}$, and often violate semantic consistency.

Thus UFA provides a provably better approximation of open space:
$\mathrm{Cov}\_{\text{UFA}}(\mathcal{O})
\supset
\mathrm{Cov}\_{\text{image-aug}}(\mathcal{O}),$
where $\mathrm{Cov}(\cdot)$ denotes coverage measure.
% \subsubsection{Calibrated Energy Separation.}
% Let $E(\mathbf{z})$ denote free energy.  
% For any OOD point $\mathbf{z}\in\mathcal{O}$, UFA enforces \[E(\mathbf{z}) \ge m\] and further constraints predictive entropy to be large:\begin{equation}
% H(\mathbf{z}) \ge H_{\min}. 
% \end{equation}Combining the two yields a Lipschitz-type calibration bound:\begin{equation}
% \max_{y} p(y|\mathbf{z})
% \le 
% \exp(-m/\tau)
% \quad\text{for all}\quad
% \mathbf{z}\in\mathcal{O},
% \end{equation}
% which prevents overconfident misclassification in open space.Prior confidence penalties lack density-awareness and cannot guarantee
% such separation, leading to false absorption.
\subsubsection{Calibrated Energy Separation.}
Let $E(\mathbf{z})$ denote free energy.
For any OOD point $\mathbf{z}\in\mathcal{O}$, UFA enforces $E(\mathbf{z}) \ge m$ and further constrains predictive entropy to be large:
$H(\mathbf{z}) \ge H_{\min}.$
Combining the two yields a Lipschitz-type calibration bound:
$\max_{y} p(y|\mathbf{z})
\le \exp(-m/\tau) ; \forall 
\mathbf{z}\in\mathcal{O}$
which prevents overconfident misclassification in open space. Prior confidence penalties lack density-awareness and cannot guarantee
such separation, leading to false absorption.

\subsection{\textbf{Training and Inference}}
We train {\ourmodel} end-to-end, minimizing a composite loss that jointly enforces discriminative, domain-robust, and uncertainty-calibrated representations:
{\small
\begin{equation}
\label{eq:total_loss}
\mathcal{L}_{\mathrm{total}}
= \lambda_{\mathrm{NCE}}\mathcal{L}_{\mathrm{InfoNCE}}
+ \lambda_{\mathrm{Scon}}\mathcal{L}_{\mathrm{SCon}}
+ \lambda_{\mathrm{ce}}\mathcal{L}_{\mathrm{CE}}
+ \mathcal{L}_{\mathrm{UFA}},
\end{equation}
}
where $\lambda_{\mathrm{NCE}},\lambda_{\mathrm{Scon}}, \lambda_{\mathrm{ce}}$ balance the contributions.
The InfoNCE term $\mathcal{L}_{\text{InfoNCE}}$ aligns augmented views to build view- and domain-invariant embeddings, while the supervised contrastive loss ($\mathcal{L}_{\mathrm{SCon}}$) clusters samples from the same class and separates different ones—key for fine-grained datasets with subtle inter-class variations.
A standard cross-entropy term ($\mathcal{L}_{\mathrm{CE}}$) aligns the cosine classifier $g(\cdot)$ with the labeled source data, ensuring stable decision boundaries for known classes.
These losses, together with $\mathcal{L}_{\text{UFA}}$, yield embeddings that are compact and discriminative for known classes yet calibrated and separable for unseen ones. Additional loss function details are provided in the \textbf{Sup. Mat.}

During \textbf{testing}, we follow \cite{gcd,dg2net} and apply K-Means clustering on $\mathcal{D}_t$, using the K-estimation strategy of \cite{gcd,rongali2024cdadnetbridgingdomaingaps} to infer the number of clusters (see \textbf{Sup. Mat.})
\section{Experimental Evaluations}
\label{sec:experiments}
%%%%%%%%%%%%%%%%%%%%%%%%%%%%

\noindent\textbf{Datasets.}
We evaluate \ourmodel\ on three fine-grained benchmarks—\textbf{CUB-200-2011}~\cite{CUB_200_2011} (200 bird species), \textbf{Stanford Cars}~\cite{scars} (196 car models), and \textbf{FGVC-Aircraft}~\cite{fgvc_aircarft} (100 aircraft variants)—each exhibiting high intra-class variability and subtle inter-class cues.
As these datasets originate from a single photographic domain, they lack the stylistic diversity necessary to assess cross-domain generalization.
% To address this, we augment each dataset with two additional style domains—\texttt{painting} and \texttt{sketch}—forming a tri-domain setup for systematic evaluation.
To address this, we construct \textit{multi-domain variants}—denoted as \textbf{CUB-200-2011-MD}, \textbf{Stanford Cars-MD}, and \textbf{FGVC-Aircraft-MD}—by augmenting each dataset with two additional style domains: \texttt{painting} and \texttt{sketch}.
Each domain is alternately treated as the source, while the remaining domains serve as unseen targets under a known-to-novel class split. Specifically, we adopt splits of CUB-200-2011 (100:100), Stanford Cars (98:98), and FGVC-Aircraft (50:50), following the CD-GCD and DG evaluation protocols~\cite{rongali2024cdadnetbridgingdomaingaps,gcd}.

\noindent\textbf{Implementation Details.} 
We implement \ourmodel\ in \texttt{PyTorch} using a ViT-B/16 backbone~\cite{vit} pretrained with DINO~\cite{dino}. Following prior DG-GCD literature~\cite{dg2net,Rathore2025HiDISC}, we fine-tune only the final transformer block to maintain stability and efficiency. Input images are resized to $224{\times}224$ and augmented into two views ($n_{\text{views}}=2$) using the \texttt{imagenet\_dg\_strong} transformation suite (details in \textbf{Sup. Mat.}). These views pass through the DCPD module, where standard \texttt{[CLS]} tokens are replaced by part-aware representations and projected through a lightweight MLP to obtain $\ell_2$-normalized embeddings.

For optimization, we use Stochastic Gradient Descent (SGD) with a momentum of $0.9$, weight decay of $5\times10^{-5}$, and an initial learning rate of $0.3$ scheduled via cosine annealing. Training is conducted on a single NVIDIA A100 GPU for $101$ epochs with a batch size of $128$. We adopt the standard weighting $(\lambda_{\text{NCE}},\lambda_{\text{Scon}})=(0.65,0.35)$ from \cite{gcd,Rathore2025HiDISC}, while for UFA, we set $(\lambda_{\text{ce}},\lambda_{\text{OE}},\lambda_{\text{ENT}})=(1.0,0.5,0.5)$ using $64$ synthetic uncertainty samples per batch. To ensure fair comparison and isolate the impact of our contributions, we integrated the DCPD module into several baseline methods by replacing their native input embeddings; empirical results confirm that DCPD consistently enhances the performance of existing frameworks across all domain-specific experiments (\textit{real}, \textit{sketch}, \textit{painting}).

\subsection{\textbf{Evaluation on Fine-Grained Benchmarks}}
\label{subsec:fine_grained_eval}

\paragraph{\textbf{Implementation and Adaptation Details of Baselines}}

To ensure a fair comparison, all baselines use the \texttt{ViT-B/16} backbone pre-trained with DINO. We strictly follow the Domain Generalization (DG) protocol: models are trained only on the source domain and evaluated on unseen target domains without any access to target samples (labeled or unlabeled) during training.

\vspace{0.5em}
\noindent\textbf{GCD Baselines (ViT, GCD, SimGCD, CMS):} 
For standard Generalized Category Discovery methods (\textbf{ViT}~\cite{vit}, \textbf{GCD}~\cite{gcd}, \textbf{SimGCD}~\cite{gcd3}, and \textbf{CMS}~\cite{gcd6}), we follow the common practice of fine-tuning only the final block of the backbone. Optimization relies solely on source-domain data, while the remaining DINO pre-trained weights remain frozen.

\vspace{0.5em}
\noindent\textbf{FG-GCD Baselines (InfoSieve, SelEx , PALGCD):} 
Fine-Grained GCD models \textbf{InfoSieve}~\cite{rastegar2023learn}, \textbf{SelEx}~\cite{fgvc2} and PALGCD \cite{wang2025palGCD}, originally designed for single-domain discovery, require adaptation for the DG-GCD setting. We restrict them to source-only supervision, training their information-maximization and selective-handling mechanisms on source-domain ``Old'' classes. At inference, they are directly applied to target domains to categorize known classes and discover novel ones using the learned source representations.

\vspace{0.5em}
\noindent\textbf{FG-DG Baselines (FSDG):} \\
Fine-Grained Domain Generalization (FG-DG) models such as \textbf{FSDG}~\cite{FSDG} are designed for closed-set settings where training and testing share the same label space. To adapt FSDG to the open-world DG-GCD scenario, we train it only on the source domain with labeled ``Old'' classes. During training, FSDG disentangles features into common, specific, and confounding components via its feature structuralization (FS) framework. Since it lacks a category discovery mechanism, we modify inference by removing the final fully connected classifier, extracting structure-aware embeddings from the target domain, and applying K-Means clustering to classify ``Old'' classes and discover ``New'' ones.

% \vspace{0.5em}
% \noindent\textbf{DG-GCD Baselines (CDAD-Net, DG$^2$CD-Net, \textsc{HiDISC}):} 
% We compared our framework against state-of-the-art domain generalization methods with the following specific adaptations:
% \begin{itemize}[leftmargin=*, labelsep=5pt]
%     \item \textbf{CDAD-Net}~\cite{rongali2024cdadnetbridgingdomaingaps}: Although originally designed for cross-domain adaptation, we adapted \textsc{CDAD-Net} to the DG-GCD setting by strictly removing all access to target domain statistics and data during training.
%     \item \textbf{DG$^2$CD-Net}~\cite{dg2net} and \textbf{\textsc{HiDISC}}~\cite{Rathore2025HiDISC}: Following their official implementations, we introduced two synthetic domain variants (e.g., painting and sketch styles) to simulate domain shifts during training. 
%     \item \textbf{Synthetic Data Note:} In Table~\ref{tab:results}, methods marked with ``w/ Syn'' incorporate these supplementary synthetic domains to align with their original protocols. To provide a direct comparison with non-generative baselines, entries marked ``w/o Syn'' (specifically for \textsc{HiDISC}) denote models trained strictly on the real source domain without synthetic augmentation.
% \end{itemize}
% \vspace{0.5em}
\noindent\textbf{DG-GCD Baselines (CDAD-Net, DG$^2$CD-Net, \textsc{HiDISC}):} 
We compare our framework with state-of-the-art domain generalization methods using the following adaptations:
\begin{itemize}[leftmargin=*, labelsep=5pt]
    \item \textbf{CDAD-Net}~\cite{rongali2024cdadnetbridgingdomaingaps}: Originally proposed for cross-domain adaptation, we adapt \textsc{CDAD-Net} to the DG-GCD setting by removing all access to target domain data or statistics during training.
    \item \textbf{DG$^2$CD-Net}~\cite{dg2net} and \textbf{\textsc{HiDISC}}~\cite{Rathore2025HiDISC}: Following their official implementations, we introduce two synthetic domain variants (e.g., painting and sketch) to simulate domain shifts during training.
    \item \textbf{Synthetic Data Note:} In Table~\ref{tab:results}, methods marked ``w/ Syn'' use these synthetic domains to match their original protocols. For a direct comparison with non-generative baselines, entries marked ``w/o Syn'' (for \textsc{HiDISC}) denote models trained only on the real source domain without synthetic augmentation.
\end{itemize}

\noindent \textbf{Evaluation Metrics.} We evaluate \ourmodel\ using three standard metrics~\cite{gcd,dg2net}:  
(i) \textit{Known} accuracy—clustering within $\mathcal{Y}_{\text{known}}$,  
(ii) \textit{New} accuracy—clustering quality over $\mathcal{Y}_{\text{novel}}$, and  
(iii) \textit{All} accuracy—overall performance across both sets.  
Results are averaged over all source–target pairs. \\

\vspace{1.0em}
\noindent\textbf{Quantitative Comparison to the Literature}
\label{sec:comparison_main}

\noindent\textbf{Baseline Groupings.}  
We compare {\ourmodel} against five representative groups of baselines.  
(1) \textit{Feature-only baseline:}  
ViT-DINO features are directly clustered in the target domain without fine-tuning or domain adaptation.  
(2) \textit{Single-domain GCD and fine-grained GCD models:}  
GCD~\cite{gcd}, SimGCD~\cite{gcd3}, and CMS~\cite{gcd6} represent canonical GCD frameworks trained within a single domain, whereas fine-grained extensions such as InfoSieve~\cite{rastegar2023learn} ,PALGCD \cite{wang2025palGCD}, and SelEx~\cite{fgvc2} incorporate localized attention or part selection but lack mechanisms for domain generalization.  
(3) \textit{Fine-Grained DG methods:} 
We compare against FSDG~\cite{FSDG} , which focus on learning domain-invariant features for fine-grained categories. Because they are inherently closed-set, they lack native mechanisms for open-world discovery and rely purely on the separability of their learned representations for clustering.
(4) \textit{Cross-domain GCD methods:}  
CDAD-Net~\cite{rongali2024cdadnetbridgingdomaingaps} (a domain-adaptation variant), DG$^{2}$CD-Net~\cite{dg2net}, and \textsc{HiDISC}~\cite{Rathore2025HiDISC} are DG-GCD approaches that employ synthetic stylization, episodic replay, or hyperbolic embeddings to mitigate domain gaps.
(5) \textit{Hybrid Part-Aware and DCPD-Integrated baselines:} 
To isolate the effectiveness of our local representation learning, we integrate standard GCD architectures with the prior state-of-the-art part-discovery module (APL~\cite{cvprapl}) and compare them against those same baselines equipped directly with our proposed DCPD module.

%%%%%%%%%%%%%%%%%%%%%%%%%%%%%%%%%%%%%%%%%%%%%%%%%%%%%%%%%%%%%%%%%%%%%%%%%%%%%%%%%%%%%%%%%%%%

\begin{table*}[th]
  \centering
  \renewcommand{\arraystretch}{1.05}
  \resizebox{\textwidth}{!}{%
  \begin{tabular}{lcccc|ccc|ccc|cccc}
    \toprule
    \multirow{2}{*}{\textbf{Methods}} & \multirow{2}{*}{\textbf{Venue}} 
    & \multicolumn{3}{c|}{\centering \textbf{CUB-200-2011-MD}} 
    & \multicolumn{3}{c|}{\centering \textbf{Stanford Cars-MD}} 
    & \multicolumn{3}{c|}{\centering \textbf{FGVC-Aircraft-MD}} 
    & \multicolumn{3}{c}{\centering \textbf{Average}} \\
    \cmidrule(lr){3-5} \cmidrule(lr){6-8} \cmidrule(lr){9-11} \cmidrule(lr){12-14}
     &  & \textbf{All} & \textbf{Old} & \textbf{New} 
     & \textbf{All} & \textbf{Old} & \textbf{New} 
     & \textbf{All} & \textbf{Old} & \textbf{New} 
     & \textbf{All} & \textbf{Old} & \textbf{New} \\
    \midrule

    % --- General-purpose GCD Baselines ---
    \multicolumn{14}{c}{\textit{\textbf{
    GCD Baselines}}} \\
    \rowcolor{blue!8}\textbf{ViT}~\cite{vit} & ICLR'21 & 24.17 & 24.88 & 23.53 & 16.04 & 15.40 & 17.11 & 17.37 & 16.62 & 18.18 & 19.19 & 18.97 & 19.60 \\
    \rowcolor{blue!8}\textbf{GCD}~\cite{gcd} & CVPR'22 & 33.34 & 36.11 & 30.51 & 36.46 & 34.61 & 38.89 & 18.95 & 18.49 & 20.01 & 29.58 & 29.74 & 29.80 \\
    \rowcolor{blue!8}\textbf{SimGCD}~\cite{gcd3} & ICCV'23 & 29.61 & 32.40 & 26.63 & 40.90 & 34.61 & \underline{47.12} & 33.69 & 30.50 & 36.88 & 34.73 & 32.50 & 36.88 \\
    \rowcolor{blue!8}\textbf{CMS}~\cite{gcd6} & CVPR'24 & 7.82 & 7.63 & 8.00 & 4.79 & 4.71 & 4.90 & 7.11 & 7.17 & 7.05 & 6.57 & 6.50 & 6.65 \\
    \midrule

    % --- Fine-Grained GCD Baselines ---
    \multicolumn{14}{c}{\textit{\textbf{FG-GCD Baselines}}} \\
    \rowcolor{green!8}\textbf{InfoSieve}~\cite{rastegar2023learn} & NeurIPS'23 & 33.74 & 36.20 & 31.51 & 23.59 & 22.04 & 25.64 & 27.66 & 23.20 & 32.49 & 28.33 & 27.15 & 29.88 \\
    \rowcolor{green!8}\textbf{SelEx}~\cite{fgvc2} & ECCV'24 & 25.92 & 26.83 & 25.11 & 17.16 & 16.33 & 18.79 & 18.70 & 15.72 & 21.92 & 20.59 & 19.63 & 21.94 \\
    \rowcolor{green!8}\textbf{PALGCD}~\cite{wang2025palGCD} & AAAI'25 &29.36   &31.47   &\textbf{34.95}   & 30.55 & 24.62 & 36.43 &31.68  &31.62  &34.95  &30.53  &29.24  &35.44 \\ 
    \midrule
    % --- Fine-Grained DG Baselines ---
    \multicolumn{14}{c}{\textit{\textbf{FG-DG Baseline}}} \\
    \rowcolor{green!8}\textbf{FSDG}~\cite{FSDG} & TMM'25 & 10.89 & 11.47 & 10.35 & 14.12 & 14.63 & 13.10 & 12.15 & 13.53 & 11.65 & 12.39  & 13.21 & 11.70  \\
    % \rowcolor{green!8}\textbf{CSFG}~\cite{csfg} & AAAI'26 & & & & & & & & & &  & &  \\
    \midrule

    % --- Cross-Domain GCD Baselines ---
    \multicolumn{14}{c}{\textit{\textbf{DG-GCD Baselines}}} \\
    \rowcolor{purple!8}\textbf{CDAD-Net}~\cite{rongali2024cdadnetbridgingdomaingaps} & CVPR-W'24 & 32.24 & 34.95 & 29.38 & 32.29 & 29.05 & 35.50 & 33.63 & 27.87 & 39.39 & 32.72 & 30.62 & 34.76 \\
    \rowcolor{purple!8}\textbf{DG$^{2}$CD-Net\textsuperscript{\textdagger}}~\cite{dg2net} & CVPR'25 & 30.27 & 31.71 & 27.95 & 17.85 & 19.02 & 20.57 & 23.43 & 27.88 & 32.70 & 23.85 & 26.20 & 27.07 \\
    \midrule

    % --- Hybrid Part-Aware Baselines ---
    \multicolumn{14}{c}{\textit{\textbf{Hybrid Part-Aware Baselines}}} \\
    \rowcolor{green!8}\textbf{GCD + APL}~\cite{gcd} & CVPR'22 & 18.66 & 19.40 & 17.99 & 18.97 & 18.26 & 19.92 & 14.90 & 14.34 & 15.89 & 17.51 & 17.33 & 17.93 \\
    \rowcolor{green!8}\textbf{SimGCD + APL}~\cite{gcd3} & ICCV'23 & 26.28 & 25.20 & 27.76 & 23.88 & 22.14 & 26.18 & 30.68 & \underline{33.02} & 31.84 & 26.95 & 26.79 & 28.59 \\
    \rowcolor{green!8}\textbf{CMS + APL}~\cite{gcd6} & CVPR'24 & 6.66 & 6.48 & 6.82 & 4.60 & 4.56 & 4.65 & 6.44 & 6.40 & 6.49 & 5.90 & 5.81 & 5.99 \\
    \rowcolor{green!8}\textbf{\ourmodel{}} + \textbf{APL} & -- & \underline{34.75} & \underline{37.25} & 32.12 & 33.01 & 31.12 & 35.51 & 31.09 & 25.32 & 36.87 & 32.95 & 31.23 & 34.83 \\
    \midrule

    % --- DCPD-Integrated Baselines---
    \multicolumn{14}{c}{\textit{\textbf{DCPD-Integrated Baselines}}} \\
    \rowcolor{yellow!8}\textbf{GCD + DCPD}~\cite{gcd} & CVPR'22 & 29.89 & 31.70 & 28.22 & 36.19 & 31.76 & 40.57 & 24.22 & 19.42 & 29.02 & 30.10 & 27.63 & 32.60 \\
    \rowcolor{yellow!8}\textbf{SimGCD + DCPD}~\cite{gcd3} & ICCV'23 & 30.32 & 33.04 & 27.41 & \textbf{44.55 }& \underline{37.92} & 47.6 & \textbf{35.65} & \textbf{34.26} & \underline{41.05} & \underline{36.84} & \underline{35.07} & \underline{38.68} \\
    \rowcolor{yellow!8}\textbf{CMS + DCPD}~\cite{gcd6} & CVPR'24 & 10.37 & 10.01 & 10.88 & 7.74 & 8.31 & 6.78 & 9.46 & 8.16 & 10.49 & 9.19 & 8.83 & 9.38 \\
    
    \midrule

    % --- Hyperbolic DG-GCD Baselines ---
    % \multicolumn{14}{c}{\textit{\textbf{Hyperbolic DG-GCD Baselines}}} \\
    \rowcolor{purple!8}\textbf{\textsc{HiDISC} (w/o Syn.)}~\cite{Rathore2025HiDISC} & NeurIPS'25 & 33.68 & 35.46 & 32.08 & 27.66 & 26.38 & 29.35 & 25.31 & 20.67 & 30.34 & 28.89 & 27.51 & 30.59 \\
    \rowcolor{purple!8}\textbf{\textsc{HiDISC\textsuperscript{\textdagger}} (w/ Syn.)}~\cite{Rathore2025HiDISC} & NeurIPS'25 & 34.18 & 36.10 & \underline{32.45} & 37.80 & 36.19 & 39.94 & 31.23 & 26.79 & 36.04 & 35.94 & 34.88 & 37.34 \\
    \midrule

    % --- Our Model ---
    % \multicolumn{14}{c}{\textit{\textbf{Our Proposed Model}}} \\
    \rowcolor{cyan!10}\textbf{\ourmodel{}} (Ours) & -- & \textbf{35.32} & \textbf{38.11} & 32.35 & \underline{43.36} & \textbf{38.55} & \textbf{48.12} & \underline{35.35} & 29.30 & \textbf{41.41} & \textbf{38.01} & \textbf{35.32} & \textbf{40.63} \\
    \midrule
    \textbf{$\Delta$} & -- 
    & \textcolor{green!60!black}{+0.57} & \textcolor{green!60!black}{+0.86} & \textcolor{red!70!black}{-2.5}
    & \textcolor{red!60!black}{-1.19} & \textcolor{green!60!black}{+0.63} & \textcolor{green!60!black}{+1.00}
    & \textcolor{red!60!black}{-0.30} & \textcolor{red!70!black}{-4.96} & \textcolor{green!60!black}{+0.36}
    & \textcolor{green!60!black}{+1.17} & \textcolor{green!60!black}{+0.25} & \textcolor{green!60!black}{+1.95} \\
    \bottomrule
  \end{tabular}
  }
    \parbox{0.95\textwidth}{
  \centering
  \footnotesize
  \textsuperscript{\textdagger}Following the official implementation, these methods incorporate supplementary synthetic data.}
  % \vspace{-2pt}
  \captionsetup{font=small}
  \caption{\textbf{Performance comparison} across three fine-grained datasets averaged over all domain combinations. Baselines are grouped by category (GCD, FG-GCD, FG-DG, DG-GCD and Part Aware Baselines). Our method consistently outperforms prior works, both with and without synthetic domain augmentation. $\Delta$ quantifies our method’s gain or drop relative to the best prior baseline. (\textbf{Bold}: best, \underline{underline}: second-best).}
  \label{tab:results}
  \vspace{-18pt}
\end{table*}

%%%%%%%%%%%%%%%%%%%%%%

\begin{figure*}[ht]
    \centering
    \includegraphics[width=0.9\linewidth]{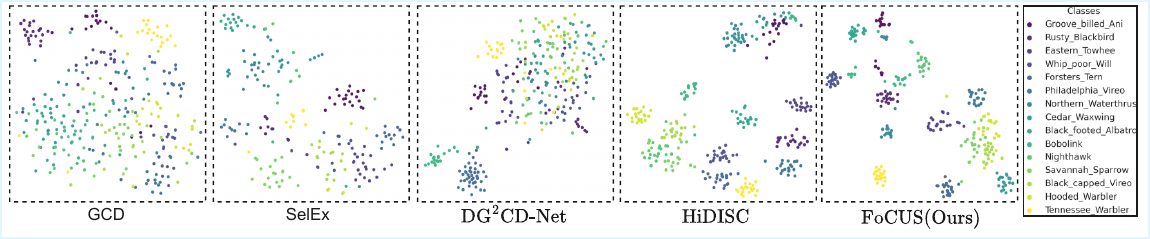}
    % \vspace{-8pt}
    \captionsetup{font=small}
    \caption{\textbf{t-SNE \cite{van2008visualizing} visualizations} of the target domain (``\texttt{Photo}") clusters, as produced by GCD \cite{gcd}, SelEx \cite{fgvc2}, DG$^{2}$CD-Net~\cite{dg2net}, \textsc{HiDISC} \cite{Rathore2025HiDISC}, and {\ourmodel} (Ours) on the CUB-200-2011-MD dataset, with ``\texttt{Sketch}" as the source domain. \ourmodel\ produces a clean and compact embedding space, clearly separating the target clusters. For visualization, we randomly select 15 classes and 20 samples per class.}
    \label{fig:tsne}
    \vspace{-8pt}
\end{figure*}

\noindent\textbf{Quantitative Comparison.}  Table~\ref{tab:results} reports results across three fine-grained benchmarks.  
Among standard GCD models, SimGCD attains the best average \textit{All} accuracy (34.73\%), surpassing GCD (29.58\%) and ViT (19.19\%), but drops to 33.69\% on FGVC-Aircraft-MD and 40.90\% on Cars-MD, revealing limited cross-domain generalization.  
Fine-grained variants like InfoSieve (28.33\%) and SelEx (20.59\%) and the recent \textsc{PALGCD} (30.53\%) improve part-level discrimination yet remain sensitive to stylistic shifts.  
Cross-domain baselines CDAD-Net (32.72\%) and DG$^{2}$CD-Net (23.85\%) use synthetic augmentation and meta-learning for better robustness but with reduced scalability.  
\textsc{HiDISC} (w/ Syn.) reaches 35.94\% average \textit{All} accuracy through hyperbolic regularization, enhancing domain coherence but compressing fine-grained manifolds and lowering separability.  
To isolate the effectiveness of localized representations, we evaluated hybrid baselines. Integrating the prior state-of-the-art \textsc{APL} module often degraded performance (e.g., dropping \textsc{SimGCD} to 26.95\%) as it overfits to domain-variant textures. Conversely, integrating our proposed \textsc{DCPD} module consistently improved its host architectures across the board (e.g., boosting \textsc{SimGCD} to a highly competitive 36.84\%), proving the superior transferability of geometry-stable priors. 
Ultimately, \ourmodel\ establishes a new state-of-the-art with an \textit{All} accuracy of 38.01\% (+1.17\% over the best prior baseline). It also records the highest \textit{New} category discovery accuracy of 40.63\% (+1.95\%), with strong, consistent gains across CUB (35.32\%), Cars (43.36\%), and Aircraft (35.35\%). These results confirm that DCPD's domain-stable features and UFA’s uncertainty calibration effectively prevent the false absorption of novel classes under severe domain shifts. Extended quantitative comparisons for Table~\ref{tab:results} are provided in the \textbf{Sup.~Mat.}
% In contrast, \ourmodel\ achieves a new state-of-the-art \textit{All} accuracy of 38.01\%, surpassing \textsc{HiDISC} by +2.07\%, SimGCD by +3.28\%, and GCD by +8.4\%.  
% It also records the highest \textit{New} accuracy of 40.63\% (+3.29\% over the next best), with consistent gains on CUB (35.32\%), Cars (43.36\%), and Aircraft (35.35\%).  
% These gains confirm that DCPD enforces geometry-consistent, domain-stable part features, while UFA’s adaptive uncertainty regularization yields coherent and well-calibrated boundaries.

Notably, \ourmodel\ attains these gains with only \textbf{5.51e15 FLOPs}, being \textbf{3--6$\times$ more efficient} than \textsc{HiDISC} (16.53e15) and GCD (33.06e15), and nearly \textbf{300$\times$ lighter} than DG$^{2}$CD-Net (1586e15).

\noindent\textbf{Qualitative Comparison.}  
The t-SNE~\cite{van2008visualizing} visualizations in Fig.~\ref{fig:tsne} corroborate these findings.  
Conventional GCD and DG-GCD models produce scattered or overlapping clusters, and \textsc{HiDISC}—though more structured—exhibits curvature-induced manifold distortions.  
By contrast, \ourmodel\ yields compact, well-separated, and semantically aligned clusters.

\subsection{\textbf{Evaluation on Coarse-Grained Benchmarks}}
\label{sec:coarse_grained_results}

\noindent\textbf{Implementation and Adaptation.} For coarse-grained evaluation on \textbf{PACS}, \textbf{Office-Home}, and \textbf{DomainNet}, we retain the same \texttt{ViT-B/16} DINO backbone and training hyperparameters used in the fine-grained experiments to highlight the inherent robustness of our framework. While prior methods such as \textsc{DG}$^2$\textsc{CD-Net} and \textsc{HiDISC} rely on extensive synthetic domain augmentation (up to 9 domains), we evaluate \ourmodel\ in two modes: \textbf{\ourmodel\ (No Synth)}, which uses no synthetic stylization, and \textbf{\ourmodel\ (2 Synth)}, which uses only two additional domains. This enables us to assess the effectiveness of our geometry-invariant DCPD and calibrated UFA modules without large-scale image-level generation.

%%%%%%%%%%%%%%%%%%%%%%%%%%%%%%%%%%%%%%%%%%%%%%%%%%%%%%%%%%%%%%%%%%%%%%%%%%%%

 \begin{table*}[th]
\centering
\resizebox{\textwidth}{!}{%
% \scriptsize
\setlength{\tabcolsep}{3pt}
\renewcommand{\arraystretch}{1.05}
\rowcolors{3}{gray!5}{white}
\begin{tabular}{lccccccccccccc}
\toprule
\textbf{Method} & \textbf{Venue} & \multicolumn{3}{c}{\textbf{PACS}} & \multicolumn{3}{c}{\textbf{Office-Home}} & \multicolumn{3}{c}{\textbf{DomainNet}} & \multicolumn{3}{c}{\textbf{Avg.}} \\
\cmidrule(lr){3-5} \cmidrule(lr){6-8} \cmidrule(lr){9-11} \cmidrule(lr){12-14}
 & & All & Old & New & All & Old & New & All & Old & New & All & Old & New \\
\midrule
\textcolor{black}{ViT} \cite{vit} & ICLR'21 & 41.98 & 50.91 & 33.16 & 26.17 & 29.13 & 21.62 & 25.35 & 26.48 & 22.41 & 31.17 & 35.51 & 25.73 \\
\textcolor{black}{GCD} \cite{gcd} & CVPR'22 & 52.28 & 62.20 & 38.39 & 52.71 & 54.19 & 50.29 & 27.41 & 27.88 & 26.13 & 44.13 & 48.09 & 38.27 \\
\textcolor{black}{SimGCD} \cite{gcd3} & ICCV'23 & 34.55 & 38.64 & 30.51 & 36.32 & 49.48 & 13.55 & 2.84  & 2.16  & 3.75  & 24.57 & 30.09 & 15.94 \\
\textcolor{black}{CMS} \cite{gcd6} & CVPR'24 & 28.95 & 28.13 & 36.80 & 10.02 & 9.66  & 10.53 & 2.33  & 2.40  & 2.17  & 13.77 & 13.40 & 16.50 \\
\textcolor{black}{SelfEx} \cite{fgvc2} & ECCV'24 & 71.82 & 73.37 & 71.55 & 50.18 & 48.59 & 52.16 & 24.78 & 24.99 & 24.21 & 48.93 & 48.98 & 49.31 \\
\textcolor{black}{CDAD-Net} \cite{rongali2024cdadnetbridgingdomaingaps} & CVPR-W'24 & 69.15 & 69.40 & 68.83 & 53.69 & 57.07 & 47.32 & 24.12 & 23.99 & 24.35 & 48.99 & 50.15 & 46.83 \\
\midrule
\textcolor{blue!60!black}{GCD+ 6 Synth} & CVPR'22 & 65.33 & 67.10 & 64.42 & 50.50 & 51.48 & 48.96 & 24.71 & 24.80 & 21.94 & 46.85 & 47.78 & 45.11 \\
\textcolor{blue!60!black}{SimGCD+ 6 Synth} & ICCV'23 & 39.76 & 43.76 & 35.97 & 35.57 & 48.58 & 12.89 & 2.71  & 1.99  & 4.14  & 26.01 & 31.44 & 17.67 \\
\textcolor{blue!60!black}{CMS+ 6 Synth} & CVPR'24 & 28.01 & 26.71 & 29.04 & 12.09 & 12.66 & 11.13 & 3.22  & 3.28  & 3.03  & 14.44 & 14.22 & 14.40 \\
\textcolor{blue!60!black}{CDAD+ 6 Synth} & CVPR-W'24 & 60.76 & 61.67 & 59.49 & 53.49 & 56.90 & 47.76 & 23.85 & 23.88 & 24.26 & 46.03 & 47.47 & 43.84 \\
\midrule
\textcolor{purple!60!black}{Hyp-GCD} \cite{Liu2025HypCD} & CVPR'25 & 65.33 & 67.11 & 64.42 & 50.13 & 49.36 & 48.08 & 22.88  & 23.74  & 25.89  & 46.12 & 46.74 & 46.13 \\
\textcolor{purple!60!black}{Hyp-SelfEx} \cite{fgvc2} & ECCV'24 & 72.44 & 74.70 & 71.20 & 52.91 & 52.65 & 52.96 & 29.30 & 30.45 & 26.37 & 51.55 & 52.60 & 50.18 \\
\midrule
\textcolor{red!70!black}{\textbf{DG$^2$CD-Net}} \cite{dg2net} (9 Synth) & CVPR'25 & 73.30 & 75.28 & 72.56 & 53.86 & 53.37 & \textbf{54.33} & 29.01 & 30.38 & 25.46 & 52.06 & 53.01 & \underline{50.78} \\
\textcolor{purple!60!black}{Hyp-DG$^2$CD-Net (9 Synth)} & CVPR'25 & 74.07 & 74.40 & \underline{73.95} & 49.40 & 50.29 & 48.03 & 22.31 & 21.52 & 24.29 & 48.59 & 48.74 & 48.76 \\

\textcolor{red!80!black}{\textbf{\textsc{HiDISC} \cite{Rathore2025HiDISC}}} (2 Synth) & NIPS'25 & \textbf{75.07} & 75.54 & \textbf{74.52} & 56.78 & 59.23 & \underline{53.21} & 30.51 & 31.40 & \underline{28.41} & 54.12 & 55.39 & \textbf{52.05} \\
\midrule
\textcolor{cyan!80!black}{\textbf{\textbf{\ourmodel} (Ours)}} (No Synth) & -- & 73.69 & \textbf{76.15} & 70.96 & \underline{57.55} & \underline{61.64} & 50.66 & \underline{32.44} & \underline{33.95} & \textbf{28.59} & \underline{54.56} & \underline{57.24} & 50.07 \\
\textcolor{cyan!80!black}{\textbf{\textbf{\ourmodel} (Ours)}} (2 Synth) & -- & \underline{74.35} & \underline{75.57} & 72.87 & \textbf{57.63} & \textbf{62.09} & 50.23 & \textbf{32.91} & \textbf{35.78} & 25.48 & \textbf{54.96} & \textbf{57.81} & 49.53 \\
\midrule
\textcolor{gray!70!black}{CDAD-Net (DA) [UB]} & CVPR-W'24 & 83.25 & 87.58 & 77.35 & 67.55 & 72.42 & 63.44 & 70.28 & 76.46 & 65.19 & 73.69 & 78.82 & 68.66 \\
\bottomrule
\end{tabular}}
\captionsetup{font=small}
\caption{
\textbf{State-of-the-Art Clustering Accuracy} (\%) on Domain Generalization Benchmarks. This table details the clustering accuracy for known (Old), novel (New), and overall (All) categories across the PACS, Office-Home, and DomainNet datasets. Our proposed \textbf{\ourmodel} consistently achieves leading performance, surpassing both synthetic domain augmentation and non-synthetic baselines. This is particularly significant as {\ourmodel} uses substantially fewer synthetic domains (down to 2 from 6/9) or even no synthetic data at all, highlighting its efficiency and effectiveness. (\textbf{Bold}: best, \underline{underline}: second best).
}
\vspace{-10pt}
\label{tab:dg-gcd-coarse}
\end{table*}

%%%%%%%%%%%%%%%%%%%%%%%%%%%%%%%%%%%%%%%%%%%%%%%%%%%%%%%%%%%%%%%%%%%%%%%%%

\noindent\textbf{Quantitative Comparison.}
While {\ourmodel} is primarily designed for fine-grained DG-GCD, its inherent robustness to domain shifts and enhanced discriminative power also yield competitive results on established coarse-grained DG-GCD benchmarks. Table \ref{tab:dg-gcd-coarse} presents a comprehensive comparison of {\ourmodel} against state-of-the-art methods on the PACS~\cite{li2017deeper}, Office-Home~\cite{officehome}, and DomainNet~\cite{peng2019moment} datasets.

Despite not being specifically optimized for coarse-grained scenarios, {\ourmodel} consistently achieves leading performance across these datasets. Notably, {\ourmodel} (2 Synth)—i.e., {\ourmodel} trained with two additional synthetic domains—achieves performance on par with state-of-the-art methods such as \textsc{HiDISC} and $DG^2CD$-Net outperforms all other methods on Office-Home for both "All" (57.63\%) and "Old" (62.09\%) categories, and secures the second-best "All" score on PACS (74.35\%). Even without any synthetic data augmentation, {\ourmodel} (No Synth) demonstrates strong capabilities, achieving the highest "Old" accuracy (76.15\%) on PACS and the second-best "All" (57.55\%) on Office-Home.

Crucially, our method often matches or surpasses leading approaches like \textsc{HiDISC} \cite{Rathore2025HiDISC} and \textbf{DG$^2$CD-Net} \cite{dg2net}, which typically rely on more extensive synthetic domain augmentation (e.g., 9 synthetic domains for \textbf{DG$^2$CD-Net} and 2 for \textsc{HiDISC}). The strong performance of {\ourmodel}, even with zero or minimal synthetic data (2 Synth), underscores the effectiveness of its core components: the Domain-Consistent Parts Discovery (DCPD) for learning geometry-invariant representations, and the Uncertainty-aware Feature Augmentation (UFA) for calibrated open-space regularization. These mechanisms, while crucial for handling subtle intra-class variations in fine-grained tasks, evidently translate well to coarse-grained settings by producing exceptionally robust, discriminative, and generalized feature embeddings. 
% This unexpected yet significant outcome highlights the broad applicability and inherent generalization capabilities of {\ourmodel} across different granularities of image data.

% \begin{table}[!ht]
% \centering
% \renewcommand{\arraystretch}{1.2}
% \setlength{\tabcolsep}{8pt}
% \resizebox{\columnwidth}{!}{%
% \begin{tabular}{lccc}
% \toprule
% \textbf{Method}               & \textbf{CUB-200-2011-MD} & \textbf{Stanford Cars-MD} & \textbf{FGVC-Aircraft-MD} \\
% \midrule
% \rowcolor{lightgrayrow} Ground Truth                  & 200  & 196  & 100 \\
% DG$^2$CD-Net         & \textcolor{red!80!black}{285} & \textcolor{red!80!black}{268} & \textcolor{red!80!black}{175} \\
% \rowcolor{lightgrayrow} \textsc{HiDISC}               & \textcolor{red!80!black}{245} & \textcolor{red!80!black}{230} & \textcolor{red!80!black}{150} \\
% \rowcolor{cyan!10} \textbf{\ourmodel~(Ours)}     & \textcolor{green!60!black}{230} & \textcolor{green!60!black}{180} & \textcolor{green!60!black}{125} \\
% \bottomrule
% \end{tabular}%
% }
% \caption{\textbf{Estimated number of clusters}. All methods deviate from the ground-truth number of categories. Our method consistently achieves the smallest error (green), while other methods show moderate (orange) or large (red) deviations.}
% \label{tab:cluster_estimation}
% \end{table}

\section{Ablation Analysis}
\label{sec:ablations}

All ablation studies are conducted on the \textbf{Stanford Cars-MD} dataset unless otherwise specified.

\subsection{\textbf{Architectural Ablations}}

\noindent \textbf{(i) Ablation on Loss Components.}  
Table~\ref{tab:ablation_loss} validates the contribution of each loss term. Starting from a 16.04\% baseline (C-0), $\mathcal{L}_{\mathrm{SCon}}$ (C-1) yields the largest boost to 28.58\%. While $\mathcal{L}_{\mathrm{InfoNCE}}$ (C-2) alone is detrimental, its synergy with $\mathcal{L}_{\mathrm{SCon}}$ (C-3) is evident. Adding $\mathcal{L}_{\mathrm{CE}}$ (C-4) raises performance to 36.22\%. The individual gains of $\mathcal{L}_{\mathrm{OE}}$ (C-5) or $\mathcal{L}_{\mathrm{ENT}}$ (C-6) are modest, but the complete model (C-8) integrating all five reaches 43.36\%. This progression highlights $\mathcal{L}_{\mathrm{SCon}}$ and $\mathcal{L}_{\mathrm{CE}}$ as core contributors, with all terms jointly enhancing performance. 

\begin{table}[!htbp]
\centering
\scriptsize % Keeps text readable but compact for side-by-side placement
\setlength{\tabcolsep}{2pt} % Slightly reduced for fit
\renewcommand{\arraystretch}{1.05}
\resizebox{0.6\linewidth}{!}{
\begin{tabular}{l ccccc ccc}
\toprule
\multirow{2}{*}{\textbf{Config.}} &
\multicolumn{5}{c}{\textbf{Loss Components}} &
\multicolumn{3}{c}{\textbf{Accuracy (\%) $\uparrow$}} \\
\cmidrule(lr){2-6} \cmidrule(lr){7-9}
 & $\mathcal{L}_{\mathrm{SCon}}$ & $\mathcal{L}_{\mathrm{InfoNCE}}$ & $\mathcal{L}_{\mathrm{CE}}$ & $\mathcal{L}_{\mathrm{OE}}$ & $\mathcal{L}_{\mathrm{ENT}}$ 
 & \textbf{All} & \textbf{Old} & \textbf{New} \\
\midrule
\rowcolor{lightgrayrow} C-0 & \ding{55} & \ding{55} & \ding{55} & \ding{55} & \ding{55} & 16.04 & 15.40 & 17.11 \\
\rowcolor{white} C-1 & \ding{51} & \ding{55} & \ding{55} & \ding{55} & \ding{55} & 28.58 & 26.92 & 30.78 \\
\rowcolor{lightgrayrow} C-2 & \ding{55} & \ding{51} & \ding{55} & \ding{55} & \ding{55} & 9.46 & 9.01 & 10.06 \\
\rowcolor{white} C-3 & \ding{51} & \ding{51} & \ding{55} & \ding{55} & \ding{55} & 31.64 & 30.32 & 33.39 \\
\rowcolor{lightgrayrow} C-4 & \ding{51} & \ding{51} & \ding{51} & \ding{55} & \ding{55} & 36.22 & 33.36 & 39.99 \\
\rowcolor{white} C-5 & \ding{51} & \ding{51} & \ding{51} & \ding{51} & \ding{55} & 36.22 & 33.73 & 39.51 \\
\rowcolor{lightgrayrow} C-6 & \ding{51} & \ding{51} & \ding{51} & \ding{55} & \ding{51} & 36.13 & 33.23 & 39.89 \\
\rowcolor{white} C-7 & \ding{51} & \ding{55} & \ding{51} & \ding{51} & \ding{51} & 37.84 & 35.34 & 41.19 \\
\midrule
\rowcolor{cyan!10} \textbf{C-8 (Full)} & \ding{51} & \ding{51} & \ding{51} & \ding{51} & \ding{51} & \textbf{43.36} & \textbf{38.55} & \textbf{48.12} \\
\bottomrule
\end{tabular}
}
\caption{\textbf{Ablation on loss components.} Evaluation on Stanford Cars-MD.}
\label{tab:ablation_loss}
\vspace{-15pt}
\end{table}

\noindent \textbf{Ablation on Feature Fusion :} Table~\ref{tab:ablation_all_first}(a) demonstrates the critical synergy of fusing global \texttt{[CLS]} features with local DCPD part representations. Our full {\ourmodel} achieves 43.36\% All. While "Only CLS" shows strong known-class performance (36.09\% Old), it lags in novel discovery (42.34\% New). "Only APL" performs poorly across the board (33.01\% All). This confirms that combining both global and local domain-consistent features is essential for optimal fine-grained DG-GCD, particularly for novel category identification.

\subsection{\textbf{Analysis of Domain-Consistent Parts Discovery (DCPD)}}

\noindent \textbf{Optimal Part Queries ($T$) and Attention Percentile ($\rho$).}  
Varying the number of learnable part prototypes ($T$) in Table~\ref{tab:ablation_all_first}(b) shows performance peaks at $T{=}16$, balancing granularity and diversity. Too few prototypes miss fine-grained cues, while too many cause redundancy. Similarly, Table~\ref{tab:ablation_all_first}(c) evaluates $\rho$, which controls the proportion of CLS$\rightarrow$patch attention retained for priors. A clear peak at $\rho = 0.30$ indicates that retaining only the top 30\% of attention responses isolates geometry-salient regions perfectly; smaller or larger percentiles either starve the priors of context or dilute them with background noise.
\begin{table*}[!ht]
\centering
\setlength{\tabcolsep}{3pt}
\renewcommand{\arraystretch}{1.05}

% ---------- SUBTABLE (a) ----------
\begin{subtable}[t]{0.32\textwidth} % Slightly increased width for better fit
\centering
\resizebox{\linewidth}{!}{% % Added resizebox
\begin{tabular}{lccc}
\toprule
\textbf{Method} & \textbf{All} & \textbf{Old} & \textbf{New} \\
\midrule
\rowcolor{lightgrayrow} Only CLS & 38.78 & 36.09 & 42.34 \\
Only DCPD & 33.01 & 31.12 & 35.51 \\
\rowcolor{cyan!10}{\ourmodel} & \textbf{43.36} & \textbf{38.55} & \textbf{48.12} \\
\bottomrule
\end{tabular}}
\caption{Fusion strategy}
\end{subtable}
\hfill
% ---------- SUBTABLE (b) ----------
\begin{subtable}[t]{0.32\textwidth}
\centering
\resizebox{\linewidth}{!}{% % Added resizebox
\begin{tabular}{lccc}
\toprule
\textbf{Parts} & \textbf{All} & \textbf{Old} & \textbf{New} \\
\midrule
\rowcolor{lightgrayrow} 12 & 35.41 & 32.83 & 38.82 \\
\rowcolor{white} 14 & 38.90 & 36.50 & 42.06 \\
\rowcolor{lightgrayrow} 18 & 35.96 & 33.55 & 39.15 \\
\rowcolor{cyan!10} 16 & \textbf{43.36} & \textbf{38.55} & \textbf{48.12} \\
\bottomrule
\end{tabular}}
\caption{Number of parts in DCPD}
\end{subtable}
\hfill
% ---------- SUBTABLE (c) ----------
\begin{subtable}[t]{0.32\textwidth}
\centering
\resizebox{\linewidth}{!}{% % Added resizebox
\begin{tabular}{c|ccc}
\toprule
$\rho$ & All & Old & New \\
\midrule
0.10 & 38.42 & 35.21 & 42.65 \\
0.20 & 40.40 & 36.96 & 44.93 \\
\rowcolor{cyan!10}0.30 & \bf 43.36 & \bf 38.55 & \bf 48.12 \\
0.40 & 38.57 & 36.17 & 41.73 \\
\bottomrule
\end{tabular}}
\caption{Prior selection percentile $\rho$.}
\end{subtable}

\caption{Ablation studies on the \textbf{Stanford Cars-MD} dataset: (a) feature fusion, (b) optimal parts $T$ in DCPD, (c) prior selection percentile.}
\label{tab:ablation_all_first}
\end{table*}

\subsection{\textbf{Analysis of Uncertainty-Aware Feature Augmentation (UFA)}}
% \noindent \textbf{(ii) Analysis of UFA.}  
As shown in Table~\ref{tab:ablation_all}(a), removing any synthetic outlier sampler sharply drops accuracy—from 43.36\% (full UFA) to 35.10\%, 32.48\%, and 31.60\% without the tail, uniform, and mix samplers, respectively—showing their complementarity. The three elements work in tandem: joint cosine–energy–entropy regularization modifies the geometry and confidence of the classifier, multi-source OOD proposals address both near-manifold and open-space uncertainty, and EMA statistics offer boundary-aware density structure. Together, they provide better separation and calibration under open-set domain shift by locally and globally constraining the energy landscape.
The uniform and mix samplers regularize low-density zones, while the tail sampler expands confident boundaries, yielding smoother transitions between known and novel classes. The margin sensitivity in Table.~\ref{tab:ablation_all}(b) shows $m{=}5$ gives the best trade-off between confidence and openness. Overall, each component works synergistically for strong domain-generalized fine-grained discovery. Compared with prior open-space regularizers (Table~\ref{tab:ufa_comparison_accuracy}), UFA surpasses all baselines, achieving 43.36\% without extra data or adversarial terms—confirming the benefit of uncertainty-guided perturbations directly in feature space for stable open-set recognition.  
\begin{table*}[th]
\centering
\scriptsize % Keeps the tables compact for a single column
\setlength{\tabcolsep}{3pt}
\renewcommand{\arraystretch}{1.1}

% ---------------- (a) UFA Components ----------------
\begin{minipage}{0.31\linewidth}
    \centering
    \resizebox{\linewidth}{!}{%
    \begin{tabular}{lccc}
    \toprule
    \textbf{UFA Comp.} & \textbf{All} & \textbf{Old} & \textbf{New} \\
    \midrule
    \rowcolor{lightgrayrow} No tail & 35.10 & 32.67 & 38.30 \\
    \rowcolor{white} No uniform & 32.48 & 30.39 & 35.25 \\
    \rowcolor{lightgrayrow} No mix & 31.60 & 29.80 & 33.97 \\
    \rowcolor{cyan!10}\ourmodel & \textbf{43.36} & \textbf{38.55} & \textbf{48.12} \\
    \bottomrule
    \end{tabular}}
    \caption*{(a) UFA components}
\end{minipage}
\hfill
% ---------------- (b) Margin Sensitivity ----------------
\begin{minipage}{0.31\linewidth}
    \centering
    \resizebox{\linewidth}{!}{%
    \begin{tabular}{lccc}
    \toprule
    \textbf{Margin} & \textbf{All} & \textbf{Old} & \textbf{New} \\
    \midrule
    \rowcolor{lightgrayrow} 2.5 & 37.26 & 33.08 & 44.57 \\
    \rowcolor{white} 7.5 & 37.59 & 33.85 & 41.30 \\
    \rowcolor{lightgrayrow} 10.0 & 36.21 & 31.95 & 40.43 \\
    \rowcolor{cyan!10} 5 & \textbf{43.36} & \textbf{38.55} & \textbf{48.12} \\
    \bottomrule
    \end{tabular}}
    \caption*{(b) Margin ($m$) sensitivity}
\end{minipage}
\hfill
% ---------------- (c) Synthetic OOD Features ----------------
\begin{minipage}{0.32\linewidth}
    \centering
    \resizebox{\linewidth}{!}{%
    \begin{tabular}{lccc}
    \toprule
    \textbf{Syn. Feats.} & \textbf{All} & \textbf{Old} & \textbf{New} \\
    \midrule
    512 & 37.96 & 35.44 & 41.28 \\
    256 & 37.83 & 35.43 & 41.00 \\
    128 & 38.27 & 35.67 & 41.71 \\
    \rowcolor{cyan!10} 64 & \textbf{43.36} & \textbf{38.55} & \textbf{48.12} \\
    \bottomrule
    \end{tabular}}
    \caption*{(c) Synthetic OOD features}
\end{minipage}
\vspace{-8pt}
\caption{\textbf{Ablation results on the Stanford Cars-MD dataset.} (a) Impact of UFA sampler. (b) Sensitivity to Margin $m$. (c) Synthetic OOD features.}
\label{tab:ablation_all}
\end{table*}

% --- SECOND TABLE: UFA COMPARISON ---
\begin{table}[ht]
\centering
\setlength{\tabcolsep}{2pt}
\renewcommand{\arraystretch}{0.9}
\resizebox{\linewidth}{!}{
\begin{tabular}{lcccccc}
\toprule
\textbf{Method} & \textbf{Space} & \textbf{Mechanism} & \textbf{Open-Space} & \textbf{All} & \textbf{Old} & \textbf{New} \\
\midrule
MixUp~\cite{zhang2017mixup} & Feature & Linear interp. & \ding{55} & 36.85 & 34.20 & 39.12 \\
CutMix~\cite{yun2019cutmix} & Image & Patch repl. & \ding{55} & 37.42 & 34.91 & 40.58 \\
Manifold~\cite{verma2019manifold} & Feature & Manifold blend. & \ding{55} & 38.05 & 35.44 & 41.32 \\
VAT~\cite{miyato2018virtual} & Feature & Adv. perturb. & Partial & 39.88 & 36.70 & 43.21 \\
OE~\cite{hendrycks2018deep} & Ext. data & OOD superv. & \ding{51} & 41.02 & 37.91 & 45.15 \\
\rowcolor{cyan!10}
\textbf{UFA (Ours)} & \textbf{Feature} & \textbf{Uncert.-driven} & \textbf{Self-gen.} & \textbf{43.36} & \textbf{38.55} & \textbf{48.12} \\
\bottomrule
\end{tabular}
}
\caption{\textbf{UFA comparison.} Uncertainty-guided feature-space sampling versus baselines.}
\vspace{-10pt}
\label{tab:ufa_comparison_accuracy}
\end{table}

\noindent \textbf{Effect of synthetic OOD count :}
Table~\ref{tab:ablation_all}(c) varies the per-step count of synthetic OOD features \( \mathcal{Z}_{\mathrm{Syn}} \in \{64, 128, 256, 512\} \). Performance peaks at 64 (All 43.36\%, Old 38.55\%, New 48.12\%) and degrades as the count increases, suggesting compact, targeted outliers best regularize boundaries; excessive synthetics introduce noise/bias. We therefore use 64 in all main experiments.

\begin{figure}[!ht]
    % --- Left Side: Calibration Figure ---
    \begin{minipage}[b]{0.48\textwidth}
        \centering
        \includegraphics[width=\linewidth]{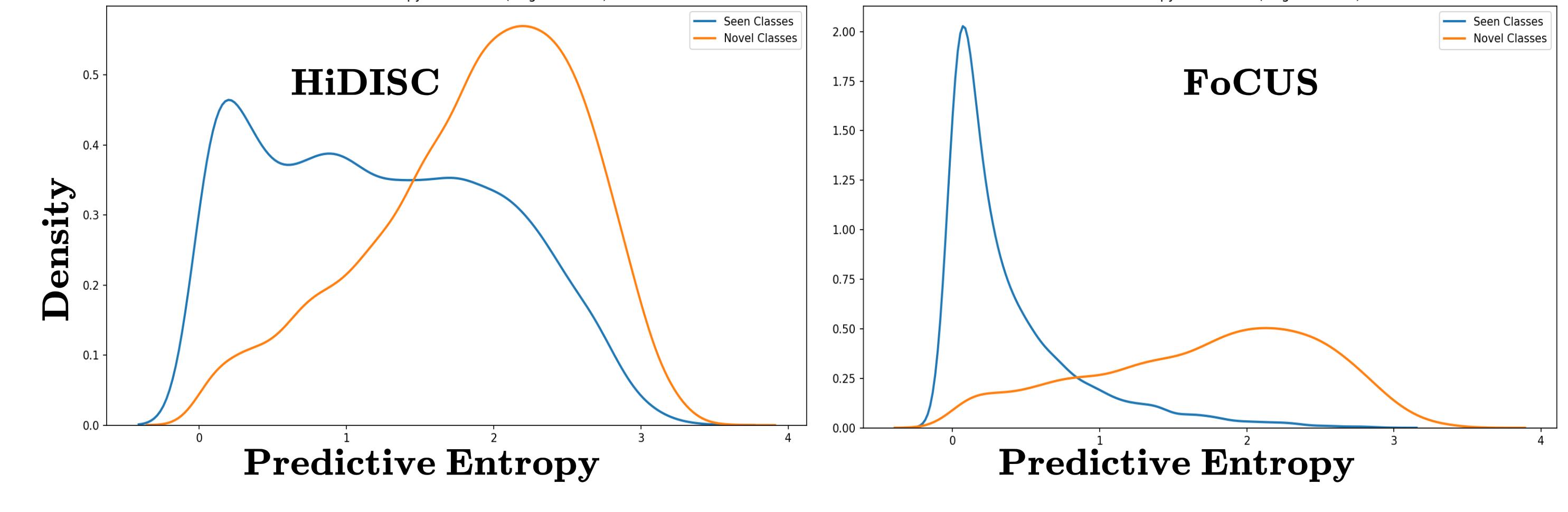}
        \caption{
            Empirical Verification. \textbf{(Left)} \textsc{HiDISC} overlaps seen (\textcolor{blue}{blue}) and novel (\textcolor{orange}{orange}) distributions (\textbf{AUROC}: 0.63). \textbf{(Right)} \textsc{FoCUS} pushes novel samples to high entropy (\textbf{AUROC}: 0.88).
        }
        \label{fig:calibration_analysis}
    \end{minipage}
    \hfill
    % --- Right Side: Data Augmentation Table ---
    \begin{minipage}[b]{0.48\textwidth}
        \centering
        \resizebox{\linewidth}{!}{%
        \begin{tabular}{lccc}
        \toprule
        \textbf{Augmentation} & \textbf{All} & \textbf{Old} & \textbf{New} \\
        \midrule
        \rowcolor{lightgrayrow} Cutout & 17.40 & 16.26 & 18.91 \\
        Rand-augment & 35.20 & 32.60 & 38.62 \\
        \rowcolor{lightgrayrow} ImageNet & 35.25 & 33.10 & 38.08 \\
        \rowcolor{cyan!10} dg\_strong & \textbf{43.36} & \textbf{38.55} & \textbf{48.12} \\
        \bottomrule
        \end{tabular}}
        \vspace{10pt} % Adds slight padding to align with figure caption baseline
        \captionof{table}{\textbf{Ablation of Data Augmentation schemes.} Evaluated on Stanford Cars-MD. The \texttt{dg\_strong} suite provides superior robustness for fine-grained discovery.}
        \label{tab:data_augmentations}
    \end{minipage}
\end{figure}

\noindent\textbf{Empirical Validation of Overconfidence.} 
A key challenge in FG-DG-GCD is the tendency of models to exhibit overconfidence in unseen regions, leading to the \emph{false absorption} of novel samples into known manifolds. As visualized in Fig.~\ref{fig:calibration_analysis}, prior state-of-the-art methods like \textsc{HiDISC} show significant overlap in entropy distributions between seen and novel categories, achieving an \textbf{Area Under the Receiver Operating Characteristic (AUROC)} curve~\cite{hendrycks2017a} of only 0.63. This confirms the baseline's susceptibility to overconfident misclassification under domain shift. 

By enforcing high-entropy predictions on synthesized OOD samples, \ourmodel\ effectively pushes novel categories into a well-calibrated, high-entropy regime (AUROC: 0.88). This yields a clear separation between known and unknown classes and demonstrates superior calibration performance.

% \begin{figure}[ht]
%     \centering
%     \includegraphics[width=\linewidth]{rebuttal_image/FG_DGCD-Page-17.png}
%     \caption{Empirical Verification. \textbf{(Left)} \textsc{HiDISC} overlaps seen (\textcolor{blue}{blue}) and novel (\textcolor{orange}{orange}) distributions (\textbf{AUROC}:\textcolor{cvprblue}{0.63}), confirming baseline overconfidence.  \textbf{(Right)} \textsc{FoCUS} pushes novel samples to high entropy (\textbf{AUROC}:\textcolor{cvprblue}{0.88}), yielding clear separation and superior calibration. On \textbf{SCARS}.}
%     \label{fig:calibration_analysis}
% \end{figure}
\vspace{-10pt}

\subsection{\textbf{Sensitivity and Robustness}}

\noindent \textbf{Ablation on augmentations :} Table~\ref{tab:data_augmentations} compares augmentation schemes for \ourmodel{}: \texttt{imagenet\_dg\_strong} yields the best All (43.36\%), surpassing baseline ImageNet (35.25\%), RandAugment (35.20\%), and Cutout (17.40\%). It consistently improves both Old and New, indicating stronger robustness to unseen domains.

% \begin{table}[ht]
% \centering
% \begin{tabular}{lccc}
% \toprule
% \textbf{Augmentation} & \textbf{All} & \textbf{Old} & \textbf{New} \\
% \midrule
% Cutout & 17.40 & 16.26 & 18.91 \\
% Rand-augment & 35.20 & 32.60 & 38.62 \\
% Imagenet & 35.25 & 33.10 & 38.08 \\
% \rowcolor{cyan!10}dg\_strong & \textbf{43.36} & \textbf{38.55} & \textbf{48.12} \\
% \bottomrule
% \end{tabular}
% \caption{Data augmentations}
% \label{tab:data_augmentations}
% \end{table}

\noindent\textbf{Hyperparameter Stability.} We evaluate \ourmodel\ across various settings for the energy margin ($m$) and the number of part queries. While performance drops occur at extreme values (e.g., excessively large margins causing over-smoothing), the model is remarkably stable near its defaults. Across three seeds using the \textbf{same} hyperparameters for all datasets, \ourmodel\ maintains an "All" metric standard deviation of $\approx 0.45$, indicating robust performance that is not overly dependent on dataset-specific tuning.

\vspace{-.15cm}
\section{Conclusions}
\label{sec:conclusions}
We introduced the challenging task of FG-DG-GCD and built the first benchmark with identity-preserving synthetic domains for systematic evaluation under stylistic shifts.
Our lightweight {\ourmodel} unifies Domain-Consistent Part Discovery with Uncertainty-aware Feature Augmentation to achieve geometry-stable, calibrated, and domain-invariant representations.
Without synthetic or episodic training, it establishes a strong baseline, showing that geometric priors markedly improve fine-grained generalization.
\textbf{Future work} will explore multi-source setups, richer domain variations, and unified architectures for scalable category discovery under real-world conditions.

% \section*{Declarations}
% \subsection*{Data Availability}
% The datasets generated and analyzed during the current study (the identity-preserving painting and sketch domains for CUB-200-2011, Stanford Cars, and FGVC-Aircraft) are available from the corresponding author on reasonable request. Upon publication, all curated benchmarks, trained model weights, and the complete \ourmodel\ codebase will be made publicly available in a dedicated GitHub repository.
\onecolumn
\centerline{\textbf{\LARGE \ourmodel: Supplementary Material}}
\section{Supplementary Material Overview}
\label{sec:nav_overview}
This supplementary document provides additional technical details, experimental results to complement the main paper. Specifically, it includes:

\begin{itemize}
    \item Section~\ref{sec:visuals}: Visualizations for the Three benchmarks
    \item Section~\ref{supp:comparison_apl}: Technical Comparison: APL vs. DCPD
    \item Section~\ref{sec:attention}: Analysis of Cross-Domain Attention Stability.
    \item Section~\ref{sec:algorithm} Detailed Training Algorithm
    \item Section~\ref{sec:loss_functions} Loss Functions Details
    \item Section~\ref{subsec:cluster_est} Cluster Estimation Robustness
    \item Section~\ref{sec:domain_results}: Results on various domains
    \item Section~\ref{subsec:aug_strategies}: Augmentation Strategies for Domain Generalization
    \item Section~\ref{sec:convergence}: Analyzing the Training Convergence
\end{itemize}

\section{Sample visualizations for the Three benchmarks}
\label{sec:visuals}

\begin{figure}[ht]
    \centering
    \includegraphics[width=\linewidth]{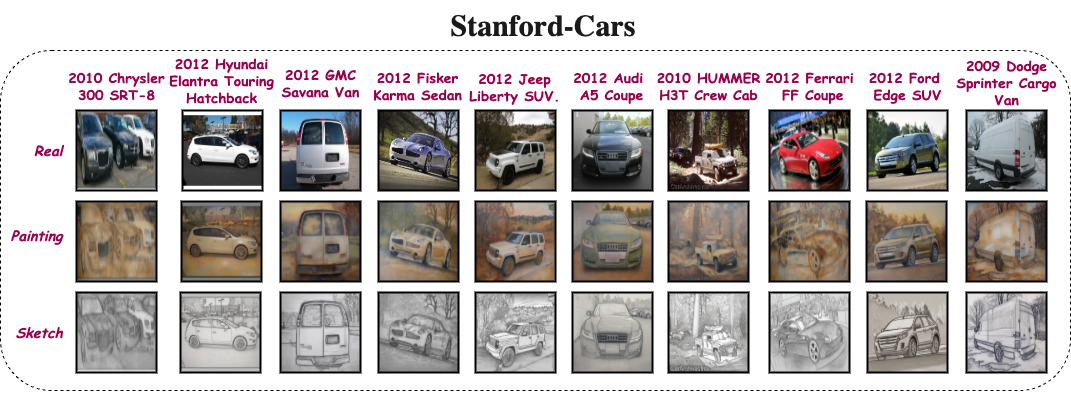}
    \vspace{-10pt}
    \caption{
    Sample images from the \textbf{Stanford Cars-MD} benchmark, illustrating diversity in make, model, and viewpoint. The benchmark contains $(16{,}185 \times 3)$ images over 196 car categories across the \texttt{real}, \texttt{painting}, and \texttt{sketch} domains.
    }
    \label{fig:SCARS_sample_dataset_1}
\end{figure}

\begin{figure}[ht]
    \centering
    \includegraphics[width=\linewidth]{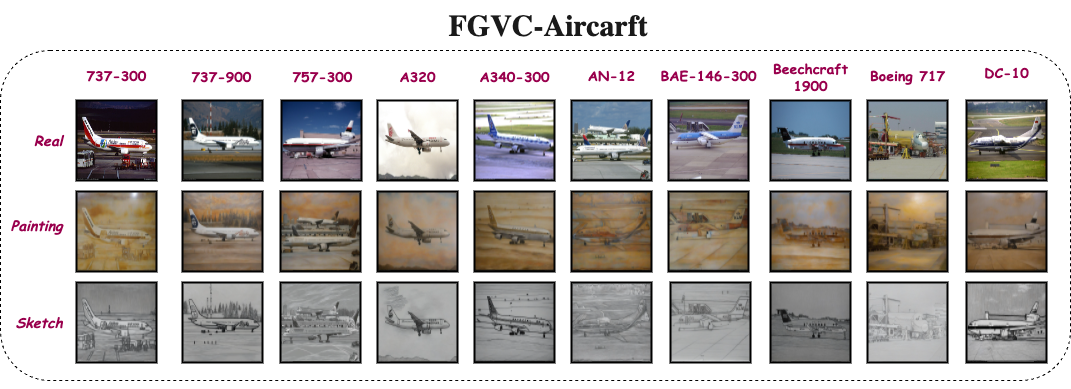}
    \vspace{-10pt}
    \caption{
    Example images from the \textbf{FGVC-Aircraft-MD} benchmark, showing variation across aircraft types, manufacturers, and liveries. The benchmark includes $(10{,}000 \times 3)$ images spanning 100 aircraft variants, each characterized by subtle structural and texture cues.
    }
    \label{fig:FGVC_Aircraft_sample_dataset}
\end{figure}

\begin{figure}[ht]
    \centering
    \includegraphics[width=\linewidth]{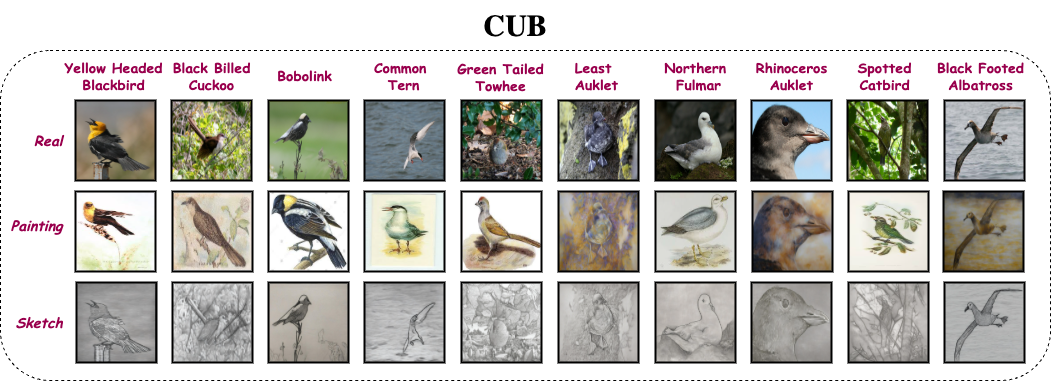}
    \caption{
    Representative samples from the \textbf{CUB-200-2011-MD} benchmark, containing $(11{,}788 \times 3)$ images from 200 bird species across three domains. Its high category count and fine visual granularity make it a particularly challenging benchmark for FG-DG-GCD.
    }
    \label{fig:CUB_sample_dataset}
\end{figure}

\clearpage
\newpage

\section{Technical Comparison: APL vs. DCPD}
\label{supp:comparison_apl}

Given that APL~\cite{cvprapl} is a primary precursor in part-based Generalized Category Discovery (GCD), we provide an explicit technical comparison to highlight the distinct innovations introduced in DCPD for the \textit{Domain Generalization} (DG) setting. While both leverage DINO-based part discovery, they differ fundamentally in their architectural philosophy and handling of distribution shifts.

\subsection{Comparison of Key Architectural Components}

The structural differences between the two frameworks are summarized in Table~\ref{tab:apl_comparison}. While APL is optimized for domain-homogeneous data, DCPD is engineered to maintain structural correspondence across drastic stylistic shifts.

\begin{table}[h]
\centering
\resizebox{\columnwidth}{!}{
\begin{tabular}{@{}lll@{}}
\toprule
\textbf{Feature} & \textbf{APL~\cite{cvprapl}} & \textbf{DCPD (Ours)} \\ 
\midrule
\textbf{Target Task} & Standard GCD & Fine-Grained Domain-Generalized GCD \\
\textbf{Query Paradigm} & Global \& Static ($\mathbf{Q}$) & Adaptive \& Instance-Conditioned ($\mathbf{Q}_I$) \\
\textbf{Discovery Prior} & Perceptual (DINO features) & Geometric Blueprint (Thresholded Masks) \\
\textbf{Global Context} & \textbf{Discarded} (Part-Replacement) & \textbf{Preserved} (Dual-Stream Fusion) \\
\textbf{Domain Handling} & Domain-Agnostic & Explicit Domain-Aware Architectural Bias \\ 
\bottomrule
\end{tabular}
}
\caption{Architectural and functional comparison between APL and the proposed DCPD.}
\label{tab:apl_comparison}
\end{table}

\subsection{Qualitative Distinction: Why Fusion Matters}
The most significant architectural departure is our \textbf{Part-Global Fusion}. In the APL formulation, the global [\texttt{CLS}] token is entirely replaced by aggregated part features to obtain the final representation:
\begin{equation}
    \mathbf{f}_{\text{APL}} = \text{Pool}(P_{\text{part}})
\end{equation}
In contrast, DCPD recognizes that while parts provide domain-stable anchors, the global token captures essential categorical context that remains relatively domain-invariant at a high semantic level. By performing a dual-stream fusion:
\begin{equation}
    \mathbf{z} = \text{L2Norm}\big(h(\mathbf{f}_{\text{cls}}) + h(\mathbf{f}_{\text{part}})\big)
\end{equation}
DCPD ensures the model does not become ``myopic''---focusing so exclusively on individual parts that it loses the holistic identity of the object. This synergy is a critical stabilizer when specific parts are occluded or severely distorted by domain shifts (e.g., from real images to sketches).

\subsection{Sensitivity to Domain Shift}
APL’s learnable queries $\mathbf{Q}$ are updated to minimize a contrastive loss across the entire training set. In a DG setting, this causes $\mathbf{Q}$ to converge on the most prominent (often texture-based) features of the source domain. Consequently, when tested on a target domain with different stylistic attributes, these static queries fail to find meaningful correspondence.

DCPD’s \textbf{Self-Attention Priors} (Eq. 2-3 in main text) act as a ``geometric filter.'' By thresholding the top-30\% of activations, we force the model to attend to high-energy structural regions such as edges and junctions rather than low-energy texture regions. This ensures that the cross-attention mechanism in the subsequent stage is grounded in a stable topology, yielding representations that are inherently more resilient to domain stylization.

\subsection{Design Rationale and Distinctness from APL.}
Unlike existing part-discovery frameworks such as APL~\cite{cvprapl} that rely on static latent embeddings, \ourmodel\ utilizes \textbf{image-conditioned part queries} attending over geometry-stable ViT priors. 

\begin{wraptable}[7]{r}{0.40\columnwidth} % Increased width slightly for readability
\vspace{-1.5em}
\centering
\scriptsize
\setlength{\tabcolsep}{3.5pt}
\renewcommand{\arraystretch}{1.05}
\begin{tabular}{@{}lcc@{}}
\toprule
\textbf{Variant} & \textbf{All} ($\uparrow$) & \textbf{PUE} ($\downarrow$)\\
\midrule
\rowcolor{lightgrayrow} APL~\cite{cvprapl} & 34.70 & 1.95 \\
\textsc{DCPD} & \textbf{43.36} & \textbf{1.72} \\
\rowcolor{white} + Div. Loss & 43.19 & 1.83 \\
\bottomrule
\end{tabular}
\captionof{table}{DCPD vs. APL.}
\label{tab:pue_ablation}
\vspace{-1.0em}
\end{wraptable}

To quantify the efficiency of our part assignment, we introduce \textbf{Parts Usage Entropy (PUE)}, defined as the Shannon entropy of the patch-to-part allocation distribution. A lower PUE indicates a sharper, more exclusive assignment where patches are assigned to distinct, semantically meaningful parts rather than being "smeared" across multiple queries. As shown in Table~\ref{tab:pue_ablation}, our annealed routing mechanism naturally achieves lower PUE (1.72) than APL (1.95). Notably, adding explicit diversity losses—a common practice in part-based learning—actually increases PUE to 1.83, causing fragmented assignments that degrade cross-domain generalization.

\newpage

\section{\textbf{Detailed Training Algorithm for \ourmodel}}
\label{sec:algorithm}
To provide a comprehensive understanding of our proposed {\ourmodel} framework, we present its detailed training procedure in Algorithm \ref{alg:focus_training}. This algorithm outlines the synergistic steps involved, from robust representation learning via Domain-Consistent Parts Discovery (DCPD) and subsequent feature fusion, to the calibrated open-space regularization enforced by Uncertainty-aware Feature Augmentation (UFA). The end-to-end training optimizes a composite loss function, ensuring discriminative power, domain invariance, and effective handling of open-set uncertainty.

\begin{algorithm}[H]
\caption{Training Procedure of {\ourmodel}}
\label{alg:focus_training}
\textbf{Input:} Source dataset $\mathcal{D}_s=\{(\mathbf{x}_i^s, y_i^s)\}$, known label set $\mathcal{Y}_{\mathrm{known}}$, pretrained ViT $E(\cdot)$, part prototypes $\mathbf{Q}$, classifier $g(\cdot)$ with weights $\{\mathbf{w}_y\}$, optimizer $\mathcal{O}$.\\
\textbf{Hyperparameters:} Gumbel temperature $\tau$, EMA decay $\alpha_{1,2}$, tail scale $\beta$, mix size $k$, batch size $B$, weights $\lambda_{\mathrm{NCE}}, \lambda_{\mathrm{SCon}}, \lambda_{\mathrm{CE}}, \lambda_{\mathrm{OE}}, \lambda_{\mathrm{ENT}}$.

\begin{algorithmic}[1]
\For{epoch $=1$ to $E$}
  \For{each minibatch $\mathcal{B}=\{(\mathbf{x}_b,y_b)\}_{b=1}^B$}
    \State \textbf{(A) DCPD Feature Extraction:} 
       Extract patch embeddings $\mathbf{F}_{\text{patch}}$, global token $\mathbf{f}_{\text{cls}}$, and attention $\mathbf{\mathcal{A}}$ from $E(\mathbf{x})$.
       Derive geometric priors $\mathbf{F}_{\text{prior}}$ using top-$\rho$ attention thresholding (Eq. \ref{eq:threshold}).
    \State \textbf{(B) Image-Conditioned Queries:} 
       $\mathbf{Q}_I = \mathrm{MHA}_{\text{cross}}(\mathbf{Q}, \mathbf{F}_{\text{prior}}, \mathbf{F}_{\text{prior}})$.
    \State \textbf{(C) Differentiable Assignment:} 
       Compute similarity $\mathbf{S} = \tilde{\mathbf{F}}_{\text{patch}} \tilde{\mathbf{Q}}_I^\top$ and routing matrix 
       $\mathcal{H} = \mathrm{GumbelSoftmax}(\mathbf{S}, \tau)$.
    \State \textbf{(D) Part-Global Fusion:} 
       Aggregate part features $\mathbf{f}_{\text{part}}$ via $\mathcal{H}$ and $\mathbf{F}_{\text{patch}}$;
       Compute fused embedding $\mathbf{z} = \mathrm{L2Norm}(h(\mathbf{f}_{\text{cls}}) + h(\mathbf{f}_{\text{part}}))$.
    \State \textbf{(E) Supervised \& Contrastive Losses:} 
       Compute $\mathcal{L}_{\mathrm{CE}}$, $\mathcal{L}_{\mathrm{InfoNCE}}$, and $\mathcal{L}_{\mathrm{SCon}}$ using $\mathbf{z}$ and labels $y_b$.
    \State \textbf{(F) Statistics Update (EMA):} 
       Update per-class statistics $(\boldsymbol{\mu}_y, \boldsymbol{\Sigma}_y)$ using batch features $\mathbf{z}_b$.
    \State \textbf{(G) UFA OOD Sampling:} 
       Generate synthetic features $\widetilde{\mathcal{Z}}$ from tail variations, between-class mixtures, and far-OOD hypersphere samples (Sec. \ref{sec:ufa}).
    \State \textbf{(H) Calibration Regularization:} 
       Compute $\mathcal{L}_{\mathrm{OE}}$ (energy) and $\mathcal{L}_{\mathrm{ENT}}$ (entropy) for all $\tilde{\mathbf{z}} \in \widetilde{\mathcal{Z}}$.
       $\mathcal{L}_{\mathrm{UFA}} = \lambda_{\mathrm{OE}}\mathcal{L}_{\mathrm{OE}} - \lambda_{\mathrm{ENT}}\mathcal{L}_{\mathrm{ENT}}$.
    \State \textbf{(I) Optimization:} 
       Update all parameters $\Theta = \{E, \mathbf{Q}, h, g\}$ by minimizing:
       $\mathcal{L}_{\text{total}} = \lambda_{\mathrm{NCE}}\mathcal{L}_{\mathrm{InfoNCE}} + \lambda_{\mathrm{SCon}}\mathcal{L}_{\mathrm{SCon}} + \lambda_{\mathrm{CE}}\mathcal{L}_{\mathrm{CE}} + \mathcal{L}_{\mathrm{UFA}}$.
  \EndFor
\EndFor
\State \textbf{Return:} Optimized model $\Theta$ and calibrated statistics $\{\boldsymbol{\mu}_y, \boldsymbol{\Sigma}_y\}$.
\end{algorithmic}
\end{algorithm}

\newpage

\newpage

\section{Further Analysis of Cross-Domain Attention Stability.}
\label{sec:attention}
In addition to the qualitative visualization shown in main paper Fig.~3, we provide a detailed quantitative study of the attention consistency achieved by our DCPD module. Specifically, we compute the pixel-wise KL divergence between the attention maps generated across different domains (Real, Painting, Sketch) for the same instance. Lower divergence indicates stronger domain-invariant localization. As reported in Table~\ref{tab:KL_div}, DCPD consistently yields the lowest cross-domain KL divergence compared to FOCUS+APL\cite{cvprapl} , confirming that it preserves geometric part cues even under severe appearance shifts. 
% We further include per-class statistics, variance plots, and additional examples illustrating that DCPD remains stable across texture-heavy domains (Painting) as well as structure-sparse domains (Sketch). These results substantiate our claim that DCPD enables reliable part discovery without synthetic augmentation or episodic training.

\begin{table}[ht]
\centering
\renewcommand{\arraystretch}{1.2}
\setlength{\tabcolsep}{10pt}
\caption{\textbf{Cross-Domain Attention Consistency (KL$\downarrow$).}
Lower values indicate more stable and domain-invariant attention maps.
We report KL divergence between domains for Real$\rightarrow$Painting,
Real$\rightarrow$Sketch, and Painting$\rightarrow$Sketch.}
\begin{tabular}{lccc}
\toprule
\textbf{Method} & \textbf{R→P} & \textbf{R→S} & \textbf{P→S} \\
\midrule
\rowcolor{lightgrayrow}FOCUS+APL~\cite{cvprapl} & 0.274 & 0.312 & 0.295 \\
% Part-Loc Baseline & 0.251 & 0.298 & 0.281 \\
% \midrule
\textbf{DCPD (Ours)} & \textbf{0.143} & \textbf{0.167} & \textbf{0.159} \\
\bottomrule
\end{tabular}
\label{tab:KL_div}
\end{table}

\newpage

\section{Loss Functions Details}
\label{sec:loss_functions}

\noindent \textbf{Total Loss.} Our overall objective combines supervised, unsupervised, and uncertainty-aware components:

\noindent\textbf{Supervised Contrastive Loss.}
To enforce class-wise consistency, we employ a supervised contrastive loss over embeddings from the previous global update \(\mathcal{F}^{g-1}\). 
Given a sample \(x\) with label \(y\) and its class prototype \(\mu^y\) (mean of embeddings for class \(y\)), the loss is:
\begin{equation}
\mathcal{L}_{\text{SCon}} = 
-\mathbb{E}_{(x,y)} 
\log 
\frac{
\exp\big(\mathrm{cos}(\mathcal{F}^{g-1}(x), \mu^y)/\tau\big)
}{
\sum_{y'\in\mathcal{Y}_s^{e_g}} 
\exp\big(\mathrm{cos}(\mathcal{F}^{g-1}(x), \mu^{y'})/\tau\big)
}.
\end{equation}
This encourages compact intra-class clusters and large inter-class margins.

\vspace{4pt}
\noindent\textbf{Unsupervised Contrastive Loss.}
To promote domain-invariant representations, we adopt an unsupervised InfoNCE-style objective. 
Each image \(x\) and its augmented view \(x^+\) form a positive pair, 
while other samples in the batch \(\mathcal{B}\) act as negatives:
\begin{equation}
\mathcal{L}_{\text{InfoNCE}} =
-\mathbb{E}_{x}
\log
\frac{
\exp\big(\mathrm{cos}(\mathcal{F}^{g-1}(x), \mathcal{F}^{g-1}(x^+))/\tau\big)
}{
\sum_{x'\in\mathcal{B}}
\exp\big(\mathrm{cos}(\mathcal{F}^{g-1}(x), \mathcal{F}^{g-1}(x'))/\tau\big)
}.
\end{equation}
This enhances local consistency and instance-level discrimination.

\vspace{4pt}
\noindent\textbf{Open-Set and Uncertainty-Aware Losses.}
We further apply:
(1) cross-entropy loss \(\mathcal{L}_{\text{CE}}\) for labeled source data;
(2) outlier energy loss \(\mathcal{L}_{\text{OE}}\) to suppress low-energy responses for synthetic outliers; and
(3) entropy regularization \(\mathcal{L}_{\text{ENT}}\) to encourage uniform predictions for uncertain samples. 
Together, these objectives promote confident predictions for known classes while maintaining high uncertainty for unknown or out-of-distribution samples.

\vspace{4pt}
This formulation ensures that our model learns class-discriminative and uncertainty-aware representations, 
leading to robust generalization across unseen domains and categories.

\noindent\textbf{Total Loss.} 
Our overall objective combines supervised, unsupervised, and uncertainty-aware components:
\begin{equation}
\mathcal{L}_{\text{total}} =
\lambda_{\text{NCE}}\mathcal{L}_{\text{InfoNCE}} +
\lambda_{\text{SCon}}\mathcal{L}_{\text{Scon}} +
\lambda_{\text{ce}}\mathcal{L}_{\text{CE}} +
\mathcal{L}_{\mathrm{UFA}},
\end{equation}
where $\mathcal{L}_{\mathrm{UFA}} = \lambda_{\text{OE}}\mathcal{L}_{\text{OE}} -
\lambda_{\text{ENT}}\mathcal{L}_{\text{ENT}}$.\\
We set \((\lambda_{\text{NCE}},\lambda_{\text{SCon}}){=}(0.65,0.35)\) following \cite{gcd,Rathore2025HiDISC}, 
and for uncertainty modeling \((\lambda_{\text{ce}},\lambda_{\text{OE}},\lambda_{\text{ENT}}){=}(1.0,0.5,0.5)\) 
with \(N_{\text{syn}}{=}64\) synthetic uncertainty samples per batch.

\vspace{4pt}

\newpage

\section{\textbf{Fine-Grained Cluster Estimation Robustness}}
\label{subsec:cluster_est}

In practical open-world settings, the number of target categories is rarely known \emph{a priori}. Table~\ref{tab:cluster_estimation} evaluates each model’s ability to estimate the number of clusters ($K$). For a fair comparison, we adopt the same $K$-estimation protocol used by \textsc{DG$^2$CD-Net}~\cite{dg2net} and \textsc{HiDISC}~\cite{Rathore2025HiDISC}. Despite this shared procedure, both baselines significantly overestimate the number of classes across all datasets, indicating susceptibility to domain shift where stylistic variations and intra-class diversity are mistakenly split into new clusters. In contrast, \ourmodel\ achieves the lowest estimation error and closely matches the ground truth on all three fine-grained benchmarks. This demonstrates that our domain-consistent part priors (DCPD) and uncertainty-aware feature alignment (UFA) effectively maintain cohesive semantic clusters under domain shift.

\begin{table}[!ht]
    \centering
    % --- Left Side: The Table Content ---
    \begin{minipage}[c]{0.52\textwidth}
        \centering
        \renewcommand{\arraystretch}{1.2}
        \setlength{\tabcolsep}{4pt} % Reduced from 8pt to fit half-page
        \resizebox{0.8\linewidth}{!}{%
        \begin{tabular}{lccc}
        \toprule
        \textbf{Method} & \textbf{CUB} & \textbf{SCARS} & \textbf{Aircraft} \\
        \midrule
        \rowcolor{lightgrayrow} Ground Truth & 200 & 196 & 100 \\
        DG$^2$CD-Net & \textcolor{red!80!black}{285} & \textcolor{red!80!black}{268} & \textcolor{red!80!black}{175} \\
        \rowcolor{lightgrayrow} \textsc{HiDISC} & \textcolor{red!80!black}{245} & \textcolor{red!80!black}{230} & \textcolor{red!80!black}{150} \\
        \rowcolor{cyan!10} \textbf{\ourmodel} & \textcolor{green!60!black}{230} & \textcolor{green!60!black}{180} & \textcolor{green!60!black}{125} \\
        \bottomrule
        \end{tabular}%
        }
    \end{minipage}
    \hfill
    % --- Right Side: The Caption ---
    \begin{minipage}[c]{0.44\textwidth}
        \caption{\textbf{Estimated number of clusters}. All methods deviate from the ground-truth number of categories. Our method consistently achieves the smallest error (green), while other methods show moderate (orange) or large (red) deviations.}
        \label{tab:cluster_estimation}
    \end{minipage}
\end{table}

\newpage 

\section{Results on various domains}
\label{sec:domain_results}

\begin{table}[!htbp]
\centering
\resizebox{0.98\textwidth}{!}{
\begin{tabular}{>{\bfseries}lccc|ccc|ccc|ccc|ccc|ccc}
\toprule
\multicolumn{19}{c}{\textbf{CUB-200-2011-MD}} \\
\midrule
 \multirow{2}{*}{\textbf{Methods}} & \multicolumn{3}{c|}{\textbf{Painting $\rightarrow$ Sketch}} & \multicolumn{3}{c|}{\textbf{Painting $\rightarrow$ Real}} & \multicolumn{3}{c|}{\textbf{Real $\rightarrow$ Sketch}} & \multicolumn{3}{c|}{\textbf{Real $\rightarrow$Painting}} & \multicolumn{3}{c|}{\textbf{Sketch$\rightarrow$ Real}} & \multicolumn{3}{c}{\textbf{Sketch $\rightarrow$ Painting}} \\
 \cmidrule(lr){2-4} \cmidrule(lr){5-7} \cmidrule(lr){8-10} \cmidrule(lr){11-13} \cmidrule(lr){14-16} \cmidrule(lr){17-19} 
 & \texttt{All} & \texttt{Old} & \texttt{New} & \texttt{All} & \texttt{Old} & \texttt{New} & \texttt{All} & \texttt{Old} & \texttt{New} & \texttt{All} & \texttt{Old} & \texttt{New} & \texttt{All} & \texttt{Old} & \texttt{New} & \texttt{All} & \texttt{Old} & \texttt{New} \\
\midrule
ViT \cite{vit}                 &37.44 & 36.27 & 38.62 & 10.48 & 11.23 & 9.82 & 28.59 & 29.73 & 27.32 & 12.27 & 12.74 & 11.86 & 25.29 & 25.32 & 25.27 & 30.96 & 34.01 & 28.28
       \\

GCD~\cite{gcd} & 47.43 & 51.14 & 43.71 & 17.64 & 18.52 & 16.86 & 35.84 & 40.44 & 31.92 & 16.32 & 17.36 & 15.40 & 33.54 & 34.28 & 32.84 & 43.83 & 45.56 & 42.30
\\

CMS~\cite{gcd6}                  & 6.47 & 6.03 & 6.85 & 5.32 & 5.34 & 5.29 & 11.30 & 11.24 & 11.35 & 5.60 & 5.42 & 5.77 & 12.13 & 11.52 & 12.70 & 6.11 & 6.21 & 6.02\\
SimGCD \cite{gcd3}               & 37.34 & 40.69 & 34.02 & 13.03 & 15.23 & 10.84 & 39.22 & 44.82 & 32.67 & 19.39 & 20.65 & 18.15 & 29.05 & 30.82 & 26.98 & 39.63 & 42.20 & 37.09

 \\
GCD+APL   & 23.53 & 24.77 & 22.44 & 9.53 & 9.28 & 9.75 & 25.29 & 25.60 & 25.00 & 11.19 & 11.81 & 10.65 & 21.62 & 22.33 & 20.95 & 20.77 & 22.64 & 19.13
   \\
CMS+APL  & 5.28 & 5.42 & 5.16 & 5.45 & 5.31 & 5.58 & 1.02 & 2.17 & 0.00& 1.02 & 2.17 & 0.00&10.81 & 10.24 & 11.35 & 5.08 & 4.95 & 5.20
      \\
SimGCD+APL & 35.04 & 36.14 & 37.01 & 15.17 & 13.07 & 17.02 & 34.65 & 31.58 & 37.57 & 17.42 & 15.63 & 19.00 & 24.74 & 24.18 & 25.27 & 30.64 & 30.58 & 30.70
 \\

CDAD-Net~\cite{rongali2024cdadnetbridgingdomaingaps} & 38.04 & 43.14 & 32.98 & 13.22 & 15.13 & 11.33 & 41.88 & 44.46 & 38.86 & 17.74 & 19.25 & 16.24 & 33.48 & 36.42 & 30.04 & 49.07 & 51.31 & 46.84
\\

InfoSieve~\cite{rastegar2023learn}  &48.72 & 53.80 & 44.23 & 16.15 & 18.31 & 14.24 & 40.23 & 42.81 & 37.78 & 15.30 & 17.12 & 13.70 & 35.38 & 37.41 & 33.46 & 46.66 & 47.77 & 45.67 
 \\

SelEX~\cite{fgvc2}  &48.73 & 49.66 & 47.91 & 15.74 & 17.18 & 14.46 & 38.77 & 38.62 & 38.92 & 16.20 & 17.26 & 15.27 & 35.07 & 36.06 & 34.12 & 1.02 & 2.17 & 0.00
 \\

DG$^2$CD-Net \cite{dg2net} &45.13 & 47.10 & 43.16 & 14.40 & 13.97 & 13.58 & 40.81 & 49.00 & 31.74 & 16.35 & 15.56 & 14.85 & 32.43 & 32.22 & 32.03 & 32.49 & 32.40 & 32.32
 \\

HiDISC~\cite{Rathore2025HiDISC} & 47.21 & 50.72 & 44.10 & 15.49 & 16.93 & 14.22 & 41.16 & 41.68 & 40.68 & 16.60 & 17.67 & 15.65 & 36.00 & 38.26 & 33.85 & 45.63 & 47.51 & 43.98
\\

\hline
{\ourmodel}(Ours) & 46.97 & 49.74 & 44.23 & 16.03 & 18.38 & 13.71 & 41.48 & 45.55 & 36.73 & 22.52 & 24.59 & 20.48 & 34.56 & 37.82 & 30.75 & 50.37 & 52.56 & 48.21
\\
\bottomrule
\end{tabular}}
\end{table}

\begin{table}[!htbp]
\centering
\resizebox{0.98\textwidth}{!}{%
\begin{tabular}{>{\bfseries}lccc|ccc|ccc|ccc|ccc|ccc}
\toprule
\multicolumn{19}{c}{\textbf{Stanford-Cars-MD}} \\ % PACS header spanning all columns
\midrule
 \multirow{2}{*}{\textbf{Methods}} & \multicolumn{3}{c|}{\textbf{Painting $\rightarrow$ Sketch}} & \multicolumn{3}{c|}{\textbf{Painting $\rightarrow$ Real}} & \multicolumn{3}{c|}{\textbf{Real $\rightarrow$ Sketch}} & \multicolumn{3}{c|}{\textbf{Real $\rightarrow$Painting}} & \multicolumn{3}{c|}{\textbf{Sketch$\rightarrow$ Real}} & \multicolumn{3}{c}{\textbf{Sketch $\rightarrow$ Painting}} \\
 \cmidrule(lr){2-4} \cmidrule(lr){5-7} \cmidrule(lr){8-10} \cmidrule(lr){11-13} \cmidrule(lr){14-16} \cmidrule(lr){17-19}
 & \texttt{All} & \texttt{Old} & \texttt{New} & \texttt{All} & \texttt{Old} & \texttt{New} & \texttt{All} & \texttt{Old} & \texttt{New} & \texttt{All} & \texttt{Old} & \texttt{New} & \texttt{All} & \texttt{Old} & \texttt{New} & \texttt{All} & \texttt{Old} & \texttt{New} \\
\midrule
ViT \cite{vit}                 &1.17 & 0.00 & 2.72 & 22.70 & 21.76 & 23.93 & 27.39 & 27.33 & 27.47 & 19.10 & 18.24 & 20.23 & 24.68 & 23.99 & 25.58 & 1.17 & 1.10 & 2.72
       \\

GCD~\cite{gcd} & 44.24 & 41.80 & 47.45 & 32.69 & 30.94 & 35.00 & 36.80 & 35.19 & 38.92 & 26.49 & 24.71 & 28.83 & 36.47 & 35.07 & 38.32 & 41.61 & 40.05 & 43.66
\\

CMS~\cite{gcd6}                  & 5.11 & 5.13 & 5.09 & 4.53 & 4.46 & 4.62 & 4.85 & 4.87 & 4.84 & 4.79 & 4.34 & 5.37 & 4.85 & 4.89 & 4.81 & 4.62 & 4.58 & 4.68\\
SimGCD \cite{gcd3}               & 48.16 & 40.53 & 55.73 & 39.18 & 32.49 & 45.80 & 42.23 & 36.32 & 48.08 & 32.51 & 27.34 & 37.62 & 38.90 & 32.97 & 44.77 & 44.41 & 38.03 & 50.73

 \\
% CDAD-Net \cite{rongali2024cdadnetbridgingdomaingaps}                 & 37.69 & 42.19 & 33.24 & 13.48 & 15.65 & 11.33 & 26.98 & 28.32 & 25.41 & 13.48 & 15.65 &
%        \\
GCD+APL   & 1.17 & 0.00 & 2.72 & 13.31 & 12.68 & 14.13 & 30.40 & 29.94 & 31.01 & 1.17 & 0.00 & 2.72 & 33.09 & 33.03 & 33.16 & 34.70 & 33.89 & 35.76
   \\
CMS+APL  & 4.57 & 4.58 & 4.55& 1.02 & 2.17 & 0.00 & 4.61 & 4.58 & 4.65 & 4.59 & 4.51 & 4.71 & 4.61 & 4.55 & 4.68& 1.02 & 2.17 & 0.00
      \\
SimGCD+APL & 29.58 & 28.16 & 31.46 & 23.67 & 21.57 & 26.43 & 22.89 & 21.07 & 25.29 & 18.20 & 15.99 & 21.12 & 25.56 & 24.98 & 26.34 & 23.37 & 21.05 & 26.43
 \\

CDAD-Net~\cite{rongali2024cdadnetbridgingdomaingaps} & 41.33 & 37.37 & 45.27 & 27.73 & 23.87 & 31.54 & 30.87 & 29.20 & 32.53 & 20.67 & 16.56 & 24.74 & 36.23 & 33.04 & 39.38 & 36.91 & 34.27 & 39.53
\\

InfoSieve~\cite{rastegar2023learn}  &29.35 & 27.97 & 31.17 & 20.34 & 17.86 & 23.62 & 25.40 & 24.66 & 26.37 & 16.76 & 15.84 & 17.96 & 22.99 & 22.00 & 24.28 & 26.72 & 23.90 & 30.45
 \\

SelEX~\cite{fgvc2}  &1.17 & 1.25 & 2.72 & 1.17 & 1.15 & 2.72 & 26.81 & 25.16 & 28.99 & 17.83 & 16.49 & 19.59 & 27.18 & 26.01 & 28.74 & 28.79 & 27.90 & 29.97
 \\

DG$^2$CD-Net \cite{dg2net} &13.73 & 14.16 & 14.73 & 16.66 & 18.50 & 20.93 & 28.00 & 29.57 & 31.65 & 18.26 & 20.18 & 22.70 & 18.02 & 18.42 & 18.94 & 12.39 & 13.29 & 14.48 
 \\

HiDISC~\cite{Rathore2025HiDISC} &36.31 & 33.34 & 40.21 & 23.59 & 22.96 & 24.41 & 29.99 & 28.72 & 31.68 & 20.04 & 19.53 & 20.71 & 26.53 & 25.07 & 28.45 & 29.50 & 28.64 & 30.64
\\

\hline
{\ourmodel}(Ours) & 50.08 & 43.72 & 56.38 & 41.44 & 35.85 & 46.99 & 43.11 & 38.07 & 48.11 & 35.14 & 29.62 & 40.61 & 43.42 & 41.56 & 45.27 & 46.94 & 42.47 & 51.37
\\
\bottomrule
\end{tabular}}
\end{table}

\begin{table}[!htbp]
\centering
\resizebox{0.98\textwidth}{!}{%
\begin{tabular}{>{\bfseries}lccc|ccc|ccc|ccc|ccc|ccc}
\toprule
\multicolumn{19}{c}{\textbf{FGVC-Aircraft-MD}} \\ % PACS header spanning all columns
\midrule
 \multirow{2}{*}{\textbf{Methods}} & \multicolumn{3}{c|}{\textbf{Painting $\rightarrow$ Sketch}} & \multicolumn{3}{c|}{\textbf{Painting $\rightarrow$ Real}} & \multicolumn{3}{c|}{\textbf{Real $\rightarrow$ Sketch}} & \multicolumn{3}{c|}{\textbf{Real $\rightarrow$Painting}} & \multicolumn{3}{c|}{\textbf{Sketch$\rightarrow$ Real}} & \multicolumn{3}{c}{\textbf{Sketch $\rightarrow$ Painting}} \\
 \cmidrule(lr){2-4} \cmidrule(lr){5-7} \cmidrule(lr){8-10} \cmidrule(lr){11-13} \cmidrule(lr){14-16} \cmidrule(lr){17-19}
 & \texttt{All} & \texttt{Old} & \texttt{New} & \texttt{All} & \texttt{Old} & \texttt{New} & \texttt{All} & \texttt{Old} & \texttt{New} & \texttt{All} & \texttt{Old} & \texttt{New} & \texttt{All} & \texttt{Old} & \texttt{New} & \texttt{All} & \texttt{Old} & \texttt{New} \\
\midrule
ViT \cite{vit}                 &17.45 & 18.22 & 16.60 & 16.10 & 14.07 & 18.29 & 21.13 & 18.97 & 23.47 & 18.41 & 18.40 & 18.41 & 14.75 & 14.88 & 14.61 & 16.37 & 15.17 & 17.67
       \\

GCD~\cite{gcd} & 37.14 & 34.20 & 40.32 & 2.01 & 3.86 & 1.10 & 31.47 & 29.18 & 33.96 & 2.01 & 3.86 & 1.15 & 2.01 & 3.86 & 1.20 & 39.03 & 35.99 & 42.32
\\

CMS~\cite{gcd6}                  & 7.49 & 7.50 & 7.49 & 7.10 & 5.94 & 8.36 & 6.80 & 6.75 & 6.87 & 6.83 & 7.09 & 6.55 & 6.98 & 7.96 & 5.93 & 7.46 & 7.79 & 7.12\\
SimGCD \cite{gcd3}               & 23.59 & 20.04 & 27.15 & 33.60 & 32.04 & 35.15 & 38.32 & 35.95 & 40.71 & 29.92 & 24.55 & 35.25 & 39.03 & 35.11 & 42.95 & 37.68 & 35.28 & 40.07

 \\
% CDAD-Net \cite{rongali2024cdadnetbridgingdomaingaps}                 & 37.69 & 42.19 & 33.24 & 13.48 & 15.65 & 11.33 & 26.98 & 28.32 & 25.41 & 13.48 & 15.65 &
%        \\
GCD+APL   & 27.04 & 22.09 & 32.40 & 19.93 & 21.40 & 18.35 & 21.19 & 18.45 & 24.16 & 17.21 & 16.38 & 18.10 & 2.01 & 3.86 & 1.05 & 2.01 & 3.86 & 1.28
   \\
CMS+APL  &6.18 & 6.23 & 6.12 & 6.74 & 6.57 & 6.93 & 6.47 & 6.69 & 6.24 & 6.38 & 5.88 & 6.93 & 6.35 & 6.23 & 6.49 & 6.53 & 6.81 & 6.24
      \\
SimGCD+APL & 34.29 & 37.77 & 36.33 & 28.84 & 32.87 & 29.34 & 30.40 & 33.56 & 30.96 & 27.52 & 30.39 & 27.78 & 30.82 & 31.31 & 32.96 & 32.19 & 32.18 & 33.65
 \\

CDAD-Net~\cite{rongali2024cdadnetbridgingdomaingaps} & 39.22 & 31.68 & 46.76 & 29.95 & 23.64 & 36.26 & 33.87 & 28.23 & 39.50 & 28.60 & 23.67 & 33.53 & 33.16 & 28.50 & 37.82 & 36.96 & 31.47 & 42.44
\\

InfoSieve~\cite{rastegar2023learn}  &31.38 & 27.39 & 35.71 & 25.15 & 20.99 & 29.65 & 29.32 & 23.82 & 35.27 & 23.74 & 18.92 & 28.96 & 27.16 & 23.13 & 31.52 & 29.23 & 24.97 & 33.83
 \\

SelEX~\cite{fgvc2}  &32.36 & 26.41 & 38.80 & 24.46 & 19.15 & 30.21 & 30.20 & 24.71 & 36.14 & 2.01 & 3.86 & 0.00 & 2.01 & 3.86 & 0.00 & 21.16 & 16.35 & 26.37
 \\

DG$^2$CD-Net \cite{dg2net} &24.51 & 27.43 & 30.59 & 21.16 & 27.55 & 34.46 & 28.84 & 32.79 & 37.08 & 20.30 & 24.40 & 28.84 & 25.78 & 32.07 & 38.89 & 20.01 & 23.05 & 26.34
 \\

HiDISC~\cite{Rathore2025HiDISC} & 29.41 & 24.74 & 34.46 & 22.36 & 18.51 & 26.53 & 28.48 & 22.15 & 35.33 & 20.32 & 17.36 & 23.53 & 24.94 & 19.38 & 30.96 & 26.38 & 21.91 & 31.21
\\

\hline
{\ourmodel}(Ours) & 37.68 & 32.40 & 42.95 & 31.68 & 26.85 & 36.50 & 39.63 & 32.01 & 47.24 & 31.06 & 26.31 & 35.81 & 35.23 & 29.70 & 40.76 & 36.84 & 28.50 & 45.17
\\
\bottomrule
\end{tabular}}
\end{table}

\newpage

\subsection{Augmentation Strategies for Domain Generalization}
\label{subsec:aug_strategies}
A central component of our training pipeline is the design of augmentation policies that expose the model to diverse and challenging input variations, thereby enforcing invariance to superficial domain-specific cues. We deliberately combine both classical appearance-based transformations and advanced style-transfer mechanisms to simulate a broad spectrum of domain shifts.

\paragraph{Strong Appearance Augmentations.}
We employ a pipeline of stochastic transformations that aggressively alter the low-level statistics of the image while preserving its semantic content. These include color jitter (hue, saturation, brightness, and contrast perturbations), random grayscale conversion, Gaussian blur, sharpness adjustment, autocontrast, solarization, and random resized cropping. Such operations induce large shifts in texture, color distribution, and edge statistics, mimicking the variability encountered across domains with different lighting, imaging devices, or post-processing artifacts.

% \paragraph{Fourier Domain Adaptation (FDA).}
% To further simulate domain shifts in frequency space, we integrate Fourier Domain Adaptation. FDA swaps the low-frequency amplitude spectrum of the input image with that of a randomly sampled style image, while retaining the original phase information. This effectively transfers global appearance properties such as color tone and illumination, while leaving object structure intact. By training on FDA-perturbed samples, the model learns to discount domain-specific biases in global statistics.

% \paragraph{Adaptive Instance Normalization (AdaIN).}
% Complementing FDA, we apply random AdaIN-based style mixing, which adjusts channel-wise statistics of content images to match those of randomly chosen style images. This allows for the injection of diverse texture patterns and stylistic cues, closely approximating domain shifts that occur due to differences in imaging environments or annotation protocols.

\paragraph{Classical Baselines.}
For comparison and stability, we also evaluate standard ImageNet-style augmentations and additional augmentation regimes such as RandAugment and Cutout. These provide a controlled baseline of moderate perturbations, ensuring that our model is not solely reliant on extreme style-transfer methods.

\paragraph{Rationale.}
Together, these augmentation strategies are designed to disentangle semantic content from domain-specific style. By systematically altering appearance while preserving object identity, the model is forced to rely on invariant cues, improving robustness to unseen target domains. This layered augmentation framework thus directly supports our objective of learning semantically meaningful and domain-agnostic representations.

\begin{figure}[ht]
    \centering
    \begin{tikzpicture}
        % Draw rounded, dashed grey rectangle around the image
        \node[
            draw=gray,          % border color
            thick,              % border thickness
            rounded corners=8pt,% rounded corner radius
            dashed,             % dashed (broken) line
            inner sep=4pt       % space between image and border
        ]{
            \includegraphics[width=0.65\textwidth]{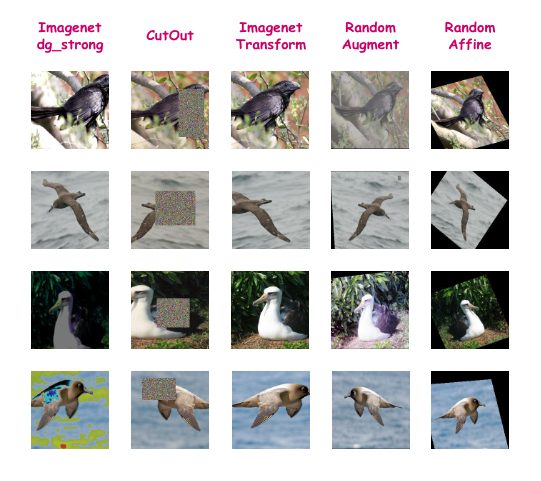}
        };
    \end{tikzpicture}
    \caption{Sample augmented images}
    \label{fig:SCARS_sample_dataset}
\end{figure}

\newpage

\section{Analyzing the Training Convergence}
\label{sec:convergence}
We illustrate the optimization dynamics on the CUB-200-2011-MD dataset in Fig.~\ref{fig:loss_convergence_total_cub}, which plots the total training loss for models trained independently on each of the three source domains. We observe stable convergence in all experiments, but with distinct optimization dynamics. The models trained on the \texttt{real} (baseline) and \texttt{sketch} domains exhibit rapid convergence, reaching a similar low-loss plateau within approximately 30 epochs. This suggests that despite the large stylistic gap of the \texttt{sketch} domain (FID: 160.18), its preservation of pure structural line-art provides a strong and unambiguous learning signal. Conversely, the \texttt{painting} domain (FID: 55.13), despite being ``closer'' in FID space, presents a far more challenging optimization task, converging significantly slower and to a substantially higher loss. This indicates that the introduction of complex artistic textures and non-photorealistic brushstrokes may act as conflicting signals or distractors, complicating the learning of subtle, fine-grained features.

\begin{figure}[!ht]
    \centering
    \begin{tikzpicture}
        \begin{axis}[
            title={\textbf{InfoNCE Loss Convergence (CUB-200-2011-MD Dataset)}},
            xlabel={Epoch},
            ylabel={Loss Value},
            width=0.9\columnwidth,
            height=6cm,
            grid=major,
            grid style=dashed,
            legend cell align={left},
            xmin=0,
            ymin=0,
            smooth,                  % Smooth curves
            no markers,              % Disable markers
            legend style={
                at={(0.5,-0.2)},     % Position below the plot
                anchor=north,        % Anchor the top of the legend
                legend columns=3     % Put all entries in one row
            }
        ]
        
        % --- Plot 1: Real Domain ---
        \addplot[smooth, thick, blue] table [x=epoch, y=info_nce, col sep=comma] {Tables/real.csv};
        \addlegendentry{Real Domain};

        % --- Plot 2: Sketch Domain ---
        \addplot[smooth, thick, red] table [x=epoch, y=info_nce, col sep=comma] {Tables/sketch.csv};
        \addlegendentry{Sketch Domain};

        % --- Plot 3: Painting Domain ---
        \addplot[smooth, thick, green!60!black] table [x=epoch, y=info_nce, col sep=comma] {Tables/painting.csv};
        \addlegendentry{Painting Domain};
        
        \end{axis}
    \end{tikzpicture}
    \caption{Comparison of \textbf{InfoNCE Loss} convergence across the three source domains on the CUB-200-2011-MD dataset.}
    \label{fig:loss_convergence_nce_cub}
\end{figure}
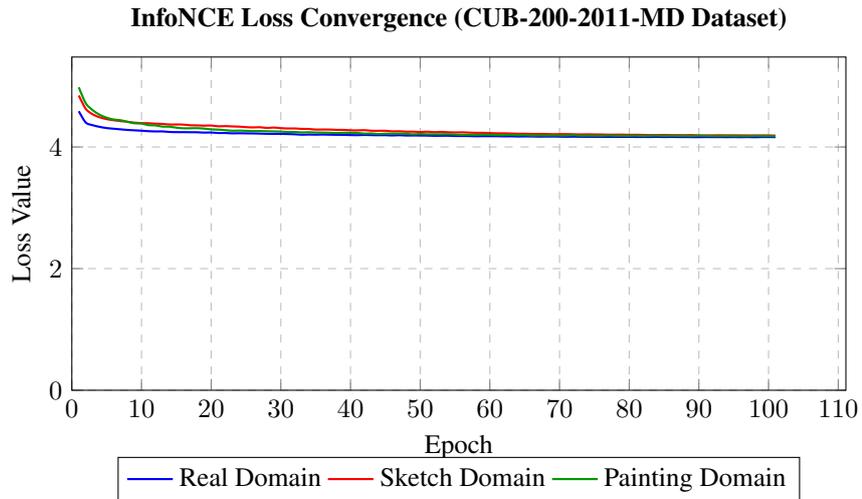

\begin{figure}[!ht]
    \centering
    \begin{tikzpicture}
        \begin{axis}[
            title={\textbf{Sup\_Con Loss Convergence (CUB-200-2011-MD Dataset)}},
            xlabel={Epoch},
            ylabel={Total Loss Value},
            width=0.9\columnwidth,
            height=6cm,
            grid=major,
            grid style=dashed,
            legend cell align={left},
            xmin=0,
            ymin=0,
            smooth,                  % Smooth curves
            no markers,              % Disable markers
            legend style={
                at={(0.5,-0.2)},     % Position below the plot
                anchor=north,        % Anchor the top of the legend
                legend columns=3     % Put all entries in one row
            }
        ]
        
        % --- Plot 1: Real Domain ---
        \addplot[smooth, thick, blue] table [x=epoch, y=sup_con, col sep=comma] {Tables/real.csv};
        \addlegendentry{Real Domain};

        % --- Plot 2: Sketch Domain ---
        \addplot[smooth, thick, red] table [x=epoch, y=sup_con, col sep=comma] {Tables/sketch.csv};
        \addlegendentry{Sketch Domain};

        % --- Plot 3: Painting Domain ---
        \addplot[smooth, thick, green!60!black] table [x=epoch, y=sup_con, col sep=comma] {Tables/painting.csv};
        \addlegendentry{Painting Domain};
        
        \end{axis}
    \end{tikzpicture}
    \caption{Comparison of \textbf{Sup\_Con Loss} convergence across the three source domains on the CUB-200-2011-MD dataset.}
    \label{fig:loss_convergence_supcon_cub}
\end{figure}

\begin{figure}[!ht]
    \centering
    \begin{tikzpicture}
        \begin{axis}[
            title={\textbf{Class Loss Convergence (CUB-200-2011-MD Dataset)}},
            xlabel={Epoch},
            ylabel={Total Loss Value},
            width=0.9\columnwidth,
            height=6cm,
            grid=major,
            grid style=dashed,
            legend cell align={left},
            xmin=0,
            ymin=0,
            smooth,                  % Smooth curves
            no markers,              % Disable markers
            legend style={
                at={(0.5,-0.2)},     % Position below the plot
                anchor=north,        % Anchor the top of the legend
                legend columns=3     % Put all entries in one row
            }
        ]
        
        % --- Plot 1: Real Domain ---
        \addplot[smooth, thick, blue] table [x=epoch, y=cls, col sep=comma] {Tables/real.csv};
        \addlegendentry{Real Domain};

        % --- Plot 2: Sketch Domain ---
        \addplot[smooth, thick, red] table [x=epoch, y=cls, col sep=comma] {Tables/sketch.csv};
        \addlegendentry{Sketch Domain};

        % --- Plot 3: Painting Domain ---
        \addplot[smooth, thick, green!60!black] table [x=epoch, y=cls, col sep=comma] {Tables/painting.csv};
        \addlegendentry{Painting Domain};
        
        \end{axis}
    \end{tikzpicture}
    \caption{Comparison of \textbf{CLS Loss} convergence across the three source domains on the CUB-200-2011-MD dataset.}
    \label{fig:loss_convergence_cls_cub}
\end{figure}

\begin{figure}[!ht]
    \centering
    \begin{tikzpicture}
        \begin{axis}[
            title={\textbf{ENT Loss Convergence (CUB-200-2011-MD Dataset)}},
            xlabel={Epoch},
            ylabel={Loss Value},
            width=0.9\columnwidth,
            height=6cm,
            grid=major,
            grid style=dashed,
            legend cell align={left},
            xmin=-5,
            ymin=-5,
            smooth,                  % Smooth curves
            no markers,              % Disable markers
            legend style={
                at={(0.5,-0.2)},     % Position below the plot
                anchor=north,        % Anchor the top of the legend
                legend columns=3     % Put all entries in one row
            }
        ]
        
        % --- Plot 1: Real Domain ---
        \addplot[smooth, thick, blue] table [x=epoch, y=ent, col sep=comma] {Tables/real.csv};
        \addlegendentry{Real Domain};

        % --- Plot 2: Sketch Domain ---
        \addplot[smooth, thick, red] table [x=epoch, y=ent, col sep=comma] {Tables/sketch.csv};
        \addlegendentry{Sketch Domain};

        % --- Plot 3: Painting Domain ---
        \addplot[smooth, thick, green!60!black] table [x=epoch, y=ent, col sep=comma] {Tables/painting.csv};
        \addlegendentry{Painting Domain};
        
        \end{axis}
    \end{tikzpicture}
    \caption{Comparison of \textbf{ENT Loss} convergence across the three source domains on the CUB-200-2011-MD dataset.}
    \label{fig:loss_convergence_ent_cub}
\end{figure}

\begin{figure}[!ht]
    \centering
    \begin{tikzpicture}
        \begin{axis}[
            title={\textbf{OE Loss Convergence (CUB-200-2011-MD Dataset)}},
            xlabel={Epoch},
            ylabel={Loss Value},
            width=0.9\columnwidth,
            height=6cm,
            grid=major,
            grid style=dashed,
            legend cell align={left},
            xmin=-5,
            ymin=-5,
            smooth,                  % Smooth curves
            no markers,              % Disable markers
            legend style={
                at={(0.5,-0.2)},     % Position below the plot
                anchor=north,        % Anchor the top of the legend
                legend columns=3     % Put all entries in one row
            }
        ]
        
        % --- Plot 1: Real Domain ---
        \addplot[smooth, thick, blue] table [x=epoch, y=oe, col sep=comma] {Tables/real.csv};
        \addlegendentry{Real Domain};

        % --- Plot 2: Sketch Domain ---
        \addplot[smooth, thick, red] table [x=epoch, y=oe, col sep=comma] {Tables/sketch.csv};
        \addlegendentry{Sketch Domain};

        % --- Plot 3: Painting Domain ---
        \addplot[smooth, thick, green!60!black] table [x=epoch, y=oe, col sep=comma] {Tables/painting.csv};
        \addlegendentry{Painting Domain};
        
        \end{axis}
    \end{tikzpicture}
    \caption{Comparison of \textbf{OE Loss} convergence across the three source domains on the CUB-200-2011-MD dataset.}
    \label{fig:loss_convergence_oe_cub}
\end{figure}

\begin{figure}[!ht]
    \centering
    \begin{tikzpicture}
        \begin{axis}[
            title={\textbf{Total Loss Convergence (CUB-200-2011-MD Dataset)}},
            xlabel={Epoch},
            ylabel={Total Loss Value},
            width=0.9\columnwidth, % Make it fit nicely in a column
            height=6cm,
            legend pos=north east, % Place legend inside the plot
            grid=major,                 % Add grid lines
            grid style=dashed,
            legend cell align={left},
            xmin=0,                     % Start x-axis at epoch 0
            ymin=0                      % Start y-axis at 0 (adjust if losses are negative)
        ]
        
        % --- Plot 1: Real Domain ---
        % 'col sep=comma' tells it to read a CSV
        % 'x=epoch' and 'y=total' match your Excel/CSV headers
        \addplot[smooth, thick, blue] table [x=epoch, y=total, col sep=comma] {Tables/cub_real.csv};
        \addlegendentry{Real Domain}; % Add its name to the legend

        % --- Plot 2: Sketch Domain ---
        \addplot[smooth, thick, red] table [x=epoch, y=total, col sep=comma] {Tables/cub_sketch.csv};
        \addlegendentry{Sketch Domain};

        % --- Plot 3: Painting Domain ---
        \addplot[smooth, thick, green!60!black] table [x=epoch, y=total, col sep=comma] {Tables/cub_painting.csv};
        \addlegendentry{Painting Domain};
        
        \end{axis}
    \end{tikzpicture}
    \caption{Comparison of \textbf{Total Loss} convergence across the three source domains on the CUB-200-2011-MD dataset.}
    \label{fig:loss_convergence_total_cub}
\end{figure}

\clearpage
\newpage

\bibliography{sn-bibliography}

\end{document}